\documentclass[a4paper,fleqn]{cas-dc}

\usepackage[numbers]{natbib}

\usepackage{amsmath,amsfonts}

\usepackage{algorithm}
\usepackage{algpseudocode}

\usepackage{array}
\usepackage[caption=false,font=normalsize,labelfont=sf,textfont=sf]{subfig}
\usepackage{textcomp}
\usepackage{stfloats}
\usepackage{url}
\usepackage{hyperref}
\usepackage{verbatim}
\usepackage{graphicx}
\usepackage{booktabs}

\usepackage{graphicx}
\usepackage{amsmath}
\usepackage{multirow}
\usepackage{algorithm}
\usepackage{algpseudocode}

\usepackage{amsfonts}

\usepackage{pgfplotstable}
\usepackage{tikz}
\usepackage{pgfplots}
\usepgfplotslibrary{groupplots}
\pgfplotsset{compat=1.18} 

\usepackage{pifont}
\newcommand{\cmark}{\ding{51}}
\newcommand{\xmark}{\ding{55}}

\usepackage{makecell}

\usepackage{xcolor}
\usepackage{pifont}

\definecolor{partialgray}{RGB}{120,120,120}
\definecolor{geomblue}{RGB}{31,119,180}

\renewcommand{\cmark}{\textcolor{green!70!black}{\ding{51}}}
\renewcommand{\xmark}{\textcolor{red!70!black}{\ding{55}}}

\usepgfplotslibrary{statistics}

\usepackage{pgfplots}
\usepackage{tikz}
\pgfplotsset{compat=1.18}

\usepackage{tabularx}

\usepackage{needspace}

\def\tsc#1{\csdef{#1}{\textsc{\lowercase{#1}}\xspace}}
\tsc{WGM}
\tsc{QE}
\tsc{EP}
\tsc{PMS}
\tsc{BEC}
\tsc{DE}

\begin{document}
\let\WriteBookmarks\relax
\def\floatpagepagefraction{1}
\def\textpagefraction{.001}
\shorttitle{UNCLE-Grasp}
\shortauthors{Malak Mansour et~al.}

\title [mode = title]{UNCLE-Grasp: A Task-Adapted Framework for Uncertainty-Aware Grasping of Leaf-Occluded Strawberries}                      



\author[1]{Malak Mansour}[orcid=0009-0005-0216-5348]
\cormark[1]
\ead{malak.mansour@mbzuai.ac.ae}

\author[1]{Ali Abouzeid}
\ead{ali.abouzeid@mbzuai.ac.ae}

\author[1]{Zezhou Sun}
\ead{zezhou.sun@mbzuai.ac.ae}

\author[1]{Qinbo Sun}
\ead{qinbo.sun@mbzuai.ac.ae}

\author[1]{Dezhen Song}
\ead{dezhen.song@mbzuai.ac.ae}

\author[1]{Abdalla Swikir}
\ead{abdalla.swikir@mbzuai.ac.ae}

\affiliation[1]{
    organization={Department of Robotics, Mohamed bin Zayed University of Artificial Intelligence},
    city={Abu Dhabi},
    country={United Arab Emirates}
}

\cortext[cor1]{Corresponding author}


\begin{abstract}
Robotic strawberry harvesting remains challenging under partial occlusion, where leaves obscure fruit geometry and makes grasp decisions based on a single shape estimate unreliable. From the same partial observation, multiple 3D completions may be plausible, and a grasp that appears feasible on one completion may fail on another. Existing uncertainty-aware grasping methods typically estimate uncertainty for pose, shape, or individual grasp candidates, but do not explicitly aggregate grasp feasibility across multiple completed object hypotheses to support an object-level attempt-or-abstain decision. This paper presents UNCLE-Grasp, a task-adapted framework that integrates learned shape completion, dropout-based variability estimation, physically grounded grasp evaluation, and risk-aware target-level grasp decisions for leaf-occluded strawberries. Monte Carlo dropout is used to generate multiple completion samples, and for each completion the retained grasp candidates are combined into a single wrench space from which a completion-level force-closure score is computed. Variability of this score across plausible strawberry completions, rather than geometric point variability alone, is then used to quantify target-level grasp uncertainty. A conservative lower confidence bound (LCB) then determines whether the system should execute a grasp or abstain from the target. The framework is evaluated in simulation and on a physical robot under increasing levels of synthetic and real leaf occlusion. At the highest simulated occlusion, UNCLE-Grasp improves success over attempted grasps from 0.780 for the strongest completed baseline to 0.870, while achieving a similar attempt rate of 0.860 versus 0.880 and increasing overall success from 0.680 to 0.740. On the physical robot, it reaches 0.800 success over attempted grasps at the highest synthetic occlusion of approximately 87\%, compared with 0.483 for the strongest completed baseline, although this improvement is accompanied by a lower attempt rate. These results show that the main benefit of UNCLE-Grasp is more reliable risk-aware execution through selective abstention, with a corresponding trade-off between grasp reliability and harvesting yield.
\end{abstract}




\maketitle

\section{Introduction}

Robotic grasping in unstructured agricultural environments remains challenging due to frequent occlusions, visual ambiguity, and partial observability of target objects \cite{zhou2022intelligent, wang2024review}. In fruit harvesting scenarios, strawberries are often partially covered by leaves, stems, or neighboring fruit \cite{gursoy2024occlusion, leaf_manipulation}, resulting in incomplete point cloud observations that can significantly distort geometric estimates. Such distortions propagate downstream to unstable or unreachable grasps and may lead to fruit damage due to poor perception and the robot's limited awareness of its own uncertainty. Moreover, harvesting is inherently sequential: grasping surrounding fruit can reduce occlusion for subsequent targets, making conservative decision-making on heavily occluded strawberries preferable to premature grasp attempts. In this setting, abstention does not need to imply permanent rejection. Difficult targets can be deferred while neighboring fruit are harvested or the viewpoint changes, and revisited once grasping becomes more feasible. This strategy is inspired by human grasping behavior, in which people naturally avoid committing to highly ambiguous grasps, adjust their viewpoint, remove surrounding obstructions, or interact with easier targets before attempting a partially hidden object. Incorporating similar uncertainty-aware and sequential reasoning may allow robotic harvesters to make safer and more reliable grasping decisions in cluttered environments.

Recent advances in learning-based grasp synthesis \cite{dexnet, gpd, contact-graspnet, mvgrasp} have demonstrated strong performance when accurate object geometry is available. These methods generate high-quality 6-DoF grasps directly from point clouds but typically assume that the observed geometry faithfully represents the underlying object. 

\begin{figure}
    \centering
    \includegraphics[width=1\linewidth]{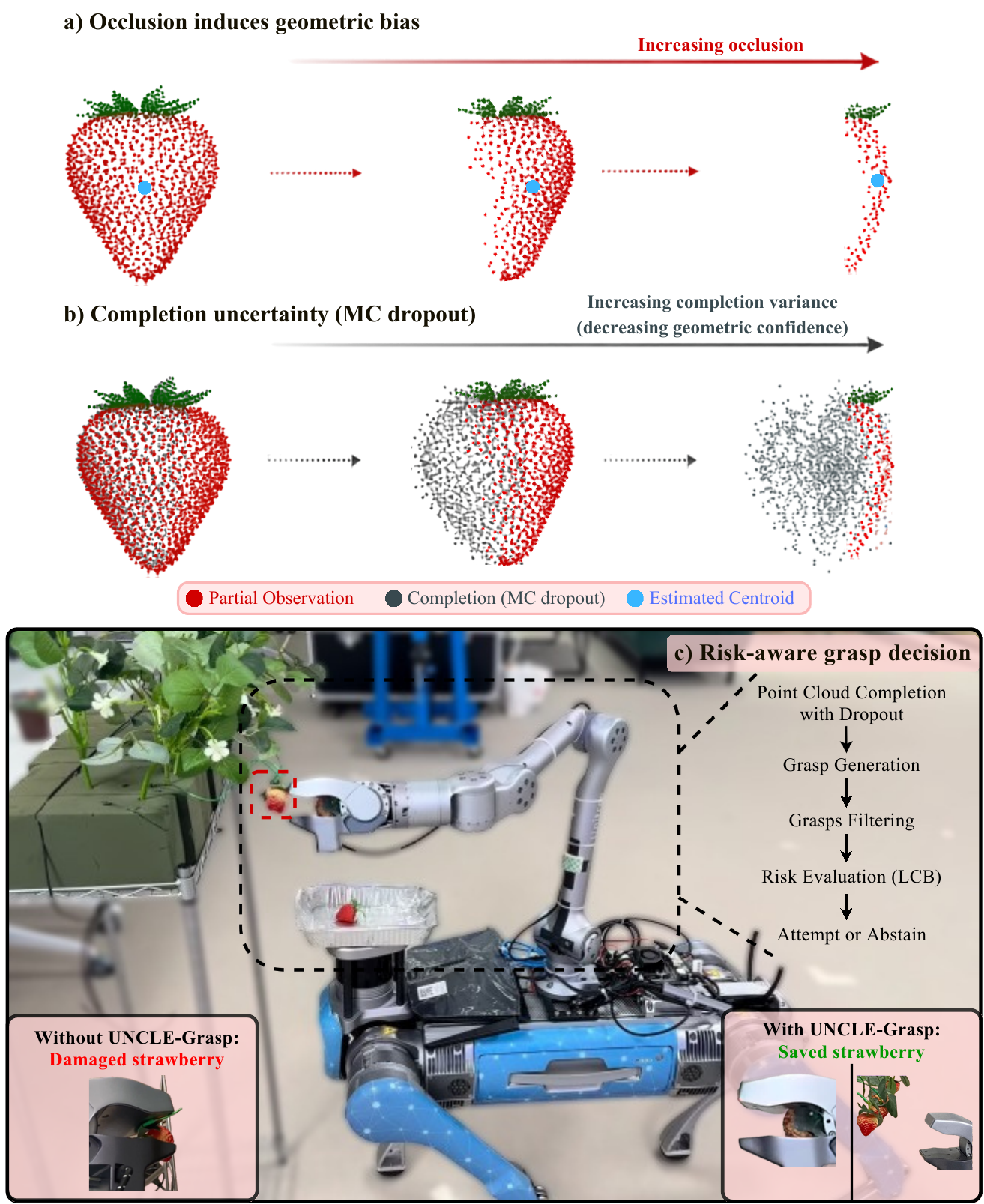}

    \caption{UNCLE-Grasp motivation. a) Increasing leaf occlusion reduces visible strawberry geometry and shifts the estimated centroid (blue dot) of the partial point cloud (red). b) MC-dropout shape completion generates multiple plausible reconstructions (gray), with increasing disagreement across samples as occlusion grows, indicating reduced geometric confidence. c) UNCLE-Grasp propagates completion uncertainty through grasp generation and filtering, evaluates object-level risk using an LCB criterion, and decides whether to attempt or abstain from grasping the target. The example outcomes illustrate how risk-aware abstention can avoid a damaging grasp and preserve the strawberry for a later attempt.}
    \label{fig:occ_levels_centroid_shift}
\end{figure}

Under partial occlusion, this assumption breaks down since perception models are known to degrade under distribution shift and incomplete observations \cite{weibel2019addressing}: grasp predictions become biased toward visible surfaces, leading to increasing centroid shifts as occlusion severity grows (Fig.~\ref{fig:occ_levels_centroid_shift}a). However, reduced geometric confidence alone is not sufficient to conclude that a grasp attempt should be avoided, since different plausible completions may still provide similar grasp support. We therefore evaluate the retained grasps for each completion through their combined grasp wrench space and compare the resulting force-closure scores $\epsilon_k$ across completion samples. If these scores remain similar, grasp feasibility is relatively stable despite variation in the reconstructed geometry; if they vary substantially, the available grasp support is sensitive to the uncertain completion and the target is considered less reliable for execution.

In addition, confidence scores produced by deterministic grasp prediction models may not adequately reflect the ambiguity introduced by missing geometry, as modern neural networks are known to yield poorly calibrated confidence estimates \cite{guo2017calibrationmodernneuralnetworks}. As a result, grasp selection can become brittle in heavily occluded scenes, motivating explicit reasoning about uncertainty in downstream grasp decisions.

A common approach to mitigating partial observability and centroid bias is point cloud completion, in which a model reconstructs the full object shape from partial observations \cite{magistri_completion_agri, 3d_shape_completion}. Although completion can reduce geometric bias, it introduces predictive uncertainty because the hidden geometry cannot always be uniquely inferred from a partial observation. Following Bayesian deep learning terminology, this uncertainty can be divided into epistemic uncertainty, which reflects uncertainty in the learned model and may increase for inputs poorly represented in the training data, and aleatoric uncertainty, which reflects measurement noise or irreducible ambiguity when multiple complete shapes are compatible with the same observation \cite{gal_gahramani_dropout,kendall_gal_uncertainty}. MC dropout approximates Bayesian inference over network weights and is therefore commonly used as a proxy for epistemic uncertainty, whereas separately estimating aleatoric uncertainty generally requires an explicit output-noise likelihood or conditional predictive distribution. In this work, variability across MC-dropout completion samples is used to approximate epistemic uncertainty, but aleatoric uncertainty is not explicitly modeled or disentangled.

Most existing grasping pipelines treat the output of completion networks as deterministic \cite{varley2017shape} and select grasps based on a single reconstructed shape, ignoring disagreement across stochastic completion predictions, in contrast to recent efforts that explicitly model completion uncertainty \cite{ville_grasp_uncertain, confidence_guided_completion}. In safety-critical manipulation tasks such as harvesting delicate fruit, this overconfidence can result in grasp failures, fruit damage, or unintended collisions with surrounding structures. Uncertainty-aware grasp evaluation has been shown to improve robustness under pose and shape ambiguity \cite{robust_grasp_pose_uncertainty_2011, phys_based_unc}. Based on this gap, we hypothesize that propagating disagreement across stochastic shape completions into grasp filtering and object-level abstention will improve success over attempted grasps under severe occlusion compared with deterministic completion-based baselines. We further expect this gain to involve a trade-off with attempt rate, which we evaluate explicitly through conditional grasp success, attempt rate, and overall harvesting success.

This paper presents an uncertainty-aware grasping pipeline for partially occluded strawberries, referred to as \emph{UNCLE-Grasp} (\textbf{UNC}ertainty-aware grasping of \textbf{LE}af-occluded strawberries). The proposed approach integrates transformer-based point cloud completion with Monte Carlo (MC) dropout \cite{gal_gahramani_dropout} to generate multiple stochastic completion predictions of the occluded fruit. For each reconstruction, candidate grasps are generated and filtered using physically motivated geometric constraints, including collision-free jaw placement, stable grasp orientation, and front-facing, kinematically feasible approach directions. We define \emph{completion-level grasp feasibility} as the force-closure support available for a strawberry under a particular stochastic completion. For each completion $k$, the retained grasp candidates are combined into a single grasp wrench space, from which one force-closure score $\epsilon_k$ is computed. Accordingly, the object-level uncertainty used for abstention is characterized by the variability of $\epsilon_k$ across plausible completions of the same strawberry, rather than by the variability of reconstructed points or individual grasp candidates alone. Geometric completion variability is used upstream to filter locally uncertain grasps and highly uncertain strawberry reconstructions, whereas the final object-level decision is based on the distribution of completion-level force-closure scores. A risk-aware Lower Confidence Bound (LCB) criterion \cite{Srinivas_2012} then determines whether a strawberry should be grasped or safely abstained by jointly accounting for the mean and variability of $\epsilon_k$ across shape completions.

In summary, this paper makes the following contributions.
\begin{itemize}

    \item An object-level decision framework that propagates uncertainty from MC-dropout shape completions (approximation of epistemic uncertainty) through grasp filtering and completion-level force-closure evaluation and uses variability in the resulting feasibility scores to determine whether grasping a partially occluded strawberry should be attempted or deferred.

    \item A task-adapted integration of global and local completion-variability filtering, strawberry-specific geometric constraints, and an LCB-style feasibility score using existing point-cloud completion and grasp-generation models without retraining them.

    \item A controlled simulation and physical-robot evaluation under synthetic and real leaf occlusions, including ablations of shape completion, geometric filtering, uncertainty-aware filtering, and object-level abstention.
\end{itemize}

\section{Related Work}
\label{sec:related_work}

Our work spans robotic grasp synthesis in cluttered scenes, 3D shape completion from partial observations, and uncertainty-aware grasp planning.
Because prior methods address different uncertainty sources and stages of the
perception-to-grasp pipeline, Table~\ref{tab:related_work_comparison}
organizes representative approaches according to their primary research
focus and summarizes how uncertainty is modeled and used. The table is
intended as a taxonomy rather than a comparison of equivalent end-to-end
systems.

\begin{table*} 
\centering
\caption{Representative uncertainty-aware grasping and shape-completion
methods grouped by primary focus. Entries summarize how uncertainty is
modeled and used without assuming equivalent end-to-end scope.}
\label{tab:related_work_comparison}
\footnotesize
\setlength{\tabcolsep}{3.5pt}
\renewcommand{\arraystretch}{1.08}

\begin{tabularx}{\textwidth}{
>{\raggedright\arraybackslash}p{2.5cm}
>{\raggedright\arraybackslash}p{3.1cm}
>{\raggedright\arraybackslash}X
>{\raggedright\arraybackslash}p{3.0cm}}
\hline
\textbf{Work}
&
\textbf{Primary Focus}
&
\textbf{Uncertainty Modeling and Use}
&
\textbf{Output / Decision Scope}
\\
\hline

\multicolumn{4}{l}{
\textit{\textbf{A. Pose and geometric uncertainty}}
}
\\
\hline

Kehoe et al.~\cite{force_closure_kehoe}
&
Push-grasp robustness
&
Monte Carlo estimation of a lower bound on force-closure probability
under uncertainty in object geometry and center of mass
&
Candidate-grasp probability bound
\\

Kim et al.~\cite{phys_based_unc}
&
Physics-based grasp evaluation
&
Dynamic grasp simulations over sampled initial object poses
&
Candidate-grasp quality estimate
\\

\hline
\multicolumn{4}{l}{
\textit{\textbf{B. Shape-completion uncertainty for grasping}}
}
\\
\hline

Lundell et al.~\cite{ville_grasp_uncertain}
&
Grasp planning over uncertain completions
&
MC-dropout completion samples are jointly evaluated using analytical
grasp metrics
&
Grasp-level selection
\\

Humt et al.~\cite{completion_predict_uncertain}
&
Prediction of uncertain completion regions
&
Predicts geometrically ambiguous reconstructed regions so that grasps
can avoid them
&
Region map and grasp avoidance
\\

Duarte et al.~\cite{measuring_uncertainty}
&
Completion-aware grasp ranking
&
MC-dropout pointwise variability is used to penalize and re-rank
grasp-quality scores
&
Grasp-candidate ranking
\\

\hline
\multicolumn{4}{l}{
\textit{\textbf{C. Uncertainty-aware grasp generation and selection}}
}
\\
\hline

Baek et al.~\cite{humanoid_uncertainty_metrics_2022}
&
Uncertainty-aware grasp selection
&
Uncertain perception- and proprioception-derived metrics are combined
through statistical ranking
&
Grasp-candidate selection
\\

Liu et al.~\cite{liu2024efficient6dof}
&
End-to-end 6-DoF grasp detection
&
Power-Spherical orientation distributions model multiple valid grasps
and uncertainty under noisy depth input
&
Grasp distribution and poses
\\

Shi et al. (vMF-Contact)~\cite{shi2025vmfcontact}
&
Uncertainty-aware contact-grasp prediction
&
Evidential vMF modeling captures aleatoric and epistemic uncertainty
in contact-conditioned grasp prediction
&
Contact-conditioned grasps
\\

\hline
\multicolumn{4}{l}{
\textit{\textbf{D. Integrated object-level decision framework}}
}
\\
\hline

\textbf{UNCLE-Grasp (Ours)}
&
\textbf{Risk-aware grasping under severe occlusion}
&
\textbf{MC completions, global/local uncertainty filtering,
force-closure evaluation, and LCB aggregation}
&
\textbf{Object-level attempt/abstain}
\\

\hline
\end{tabularx}
\end{table*}

\subsection{Grasp Synthesis from Partial Observations}

Learning-based grasp synthesis methods such as GPD~\cite{gpd} and
Dex-Net~\cite{dexnet} generate grasp candidates from point-cloud or depth
observations. More recent approaches, including Contact-GraspNet
(CGNet)~\cite{contact-graspnet} and MVGrasp~\cite{mvgrasp}, address grasp
generation in cluttered scenes from partial observations. Contact-GraspNet
generates 6-DoF grasps directly from an observed scene point cloud, while
MVGrasp constructs multiple orthographic views from an observed point cloud
for grasp prediction. These methods focus on predicting grasps from the
available observations; uncertainty over the object's unobserved geometry is
not their primary modeling target. Our approach instead performs stochastic
shape completion and uses variability across completed-shape hypotheses
during downstream grasp evaluation.

\begin{figure*} 
    \centering
\includegraphics[width=1\textwidth]{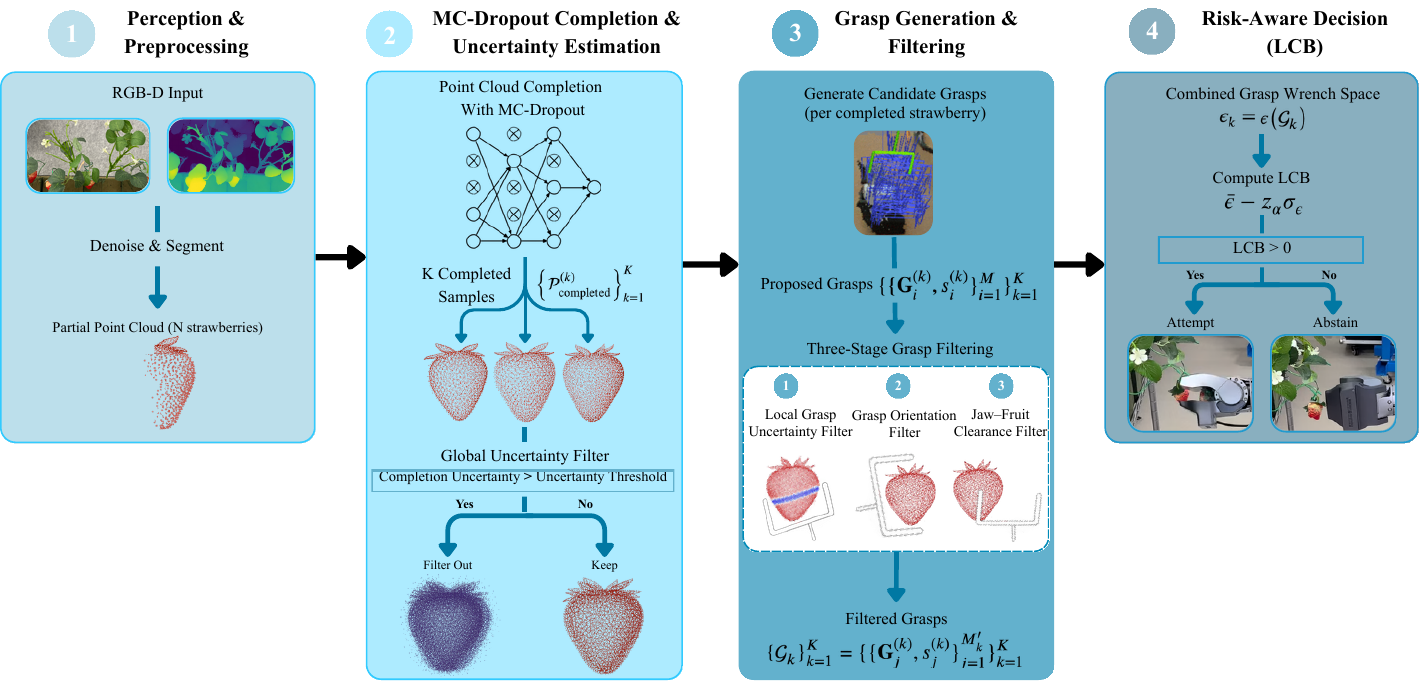}
\caption{UNCLE-Grasp pipeline.
\textbf{1) Perception \& Preprocessing:} RGB-D observations containing partially occluded strawberries are denoised and segmented to obtain an individual partial point cloud for each detected strawberry out of N strawberries.
\textbf{2) MC-Dropout Completion \& Uncertainty Estimation:} A transformer-based point cloud completion network \cite{PointAttn} reconstructs the missing geometry, with MC dropout enabled to generate $K$ stochastic completion samples $\{\mathcal{P}_{\mathrm{completed}}^{(k)}\}_{k=1}^{K}$. Variability across the completion samples provides an estimate of geometric uncertainty. Strawberries whose global completion uncertainty exceeds a predefined threshold are filtered out, while the remaining targets are retained for grasp planning.
\textbf{3) Grasp Generation \& Filtering:} A grasp-generation model \cite{contact-graspnet} generates candidate grasps for each retained completion sample. The proposed grasps are filtered using three stages: local grasp uncertainty, grasp-orientation constraints, and jaw-fruit clearance. For each completion $k$, the surviving candidates form the retained grasp set $\mathcal{G}_k$.
\textbf{4) Risk-Aware Decision (LCB):} For each completion, the wrench contributions of all grasps in $\mathcal{G}_k$ are combined into a single grasp wrench space, from which a completion-level force-closure score $\epsilon_k=\epsilon(\mathcal{G}_k)$ is computed. The resulting scores $\{\epsilon_k\}_{k=1}^{K}$ are summarized by their mean $\bar{\epsilon}$ and standard deviation $\sigma_{\epsilon}$, and the lower confidence bound $\mathrm{LCB}=\bar{\epsilon}-z_{\alpha}\sigma_{\epsilon}$ is evaluated. The system attempts to grasp the strawberry when $\mathrm{LCB}>0$ and otherwise abstains, allowing targets with inconsistent grasp feasibility across plausible completions to be deferred rather than risking an unreliable grasp attempt.}
\label{fig:pipeline}
\end{figure*}

\subsection{Shape Completion for Robotic Manipulation}

Shape completion supports robotic manipulation by estimating missing geometry from partial observations. Early learning-based approaches employed different 3D representations. For example, 3D-EPN~\cite{3D-EPN} uses a volumetric encoder-predictor network, whereas PCN~\cite{pcn} operates directly on raw point clouds.

More recent work considers confidence and uncertainty in completed geometry. Rosasco \emph{et al.}~\cite{confidence_guided_completion} propose an implicit shape representation that associates reconstructed points with confidence estimates and permits adjustment of reconstruction resolution at inference time. Humt \emph{et al.}~\cite{completion_predict_uncertain} predict regions of geometric ambiguity and evaluate the use of these predictions for avoiding uncertain grasp regions. These methods primarily provide point- or region-level information for downstream manipulation. Our method uses local completion uncertainty during grasp filtering and additionally aggregates uncertainty at the object level when deciding whether a grasp should be attempted.

In the agricultural domain, Magistri \emph{et al.}~ \cite{magistri_completion_agri} address fruit shape completion under heavy occlusion using learned geometric priors and deformable templates. Their reported framework focuses on reconstructing fruit geometry in realistic agricultural scenes. UNCLE-Grasp builds on this general motivation by considering how variability in completed fruit geometry influences downstream grasp evaluation and object-level decision making.

\subsection{Uncertainty-Aware Shape Completion and Grasp Planning}

Analytical grasp-quality measures provide a foundation for physically grounded grasp evaluation. Nguyen~\cite{force_closure_nguyen} studies the construction of force-closure grasps, while Ferrari and Canny~\cite{ferrari_canny} introduce the widely used epsilon grasp-quality metric. Kehoe \emph{et al.}~\cite{force_closure_kehoe} estimate lower bounds on force-closure probability through Monte Carlo sampling under parametric uncertainty in object geometry. Liu \emph{et al.}~\cite{differentiable_force_closure} introduce a differentiable force-closure estimator for optimizing physically stable grasps under deterministic object geometry. These works provide analytical tools that motivate the physically grounded grasp evaluation used in our framework, while considering different geometry and uncertainty assumptions.

Other work considers pose uncertainty and uncertainty in grasp-quality evaluation. Hsiao \emph{et al.}~\cite{robust_grasp_pose_uncertainty_2011} formulate manipulation under relative object-pose uncertainty, while Kim \emph{et al.}~\cite{phys_based_unc} evaluate grasp quality through physically based simulations over uncertain initial poses. Baek \emph{et al.}~\cite{humanoid_uncertainty_metrics_2022} incorporate uncertainty-aware quality metrics and sensitivity analysis into grasp-candidate selection.  SE(3)-PoseFlow~\cite{se3-poseflow} represents multimodal 6D pose distributions through flow matching on the $\mathrm{SE}(3)$ manifold and demonstrates their use in downstream manipulation. These approaches focus on pose distributions, grasp-quality uncertainty, or sensitivity to geometric perturbations. Our setting instead considers variability among multiple shapes generated by a learned completion model from the same partial observation.

More closely related work uses completion uncertainty during grasp
evaluation. Lundell \emph{et al.}~\cite{ville_grasp_uncertain} apply MC
dropout to obtain multiple stochastic shape completions and evaluate grasp
quality across the resulting samples using analytical metrics. Duarte
\emph{et al.}~\cite{measuring_uncertainty} estimate completion variability
using MC dropout and incorporate local uncertainty into grasp scoring and
candidate re-ranking. Both approaches illustrate how uncertainty in
unobserved geometry can be incorporated into candidate-level grasp
evaluation.

Complementary work models uncertainty during grasp generation. Liu \emph{et al.}~\cite{liu2024efficient6dof} represent multiple valid 6-DoF grasp orientations probabilistically and account for uncertainty associated with noisy single-view depth observations. Shi \emph{et al.}~\cite{shi2025vmfcontact} propose vMF-Contact, which combines evidential learning with probabilistic directional modeling to estimate aleatoric and epistemic uncertainty in contact-based grasp predictions. These methods incorporate uncertainty at the grasp-orientation or contact-prediction stage.

The representative methods in Table~\ref{tab:related_work_comparison} differ in their uncertainty sources, system outputs, and evaluation protocols. Accordingly, the table provides a qualitative taxonomy rather than a direct quantitative benchmark. The reviewed methods use uncertainty for pose estimation, shape completion, grasp generation, grasp evaluation, or candidate ranking. UNCLE-Grasp considers an additional object-level output: whether to attempt grasping a strawberry after aggregating physically grounded grasp feasibility across stochastic completion predictions. Here, each completion contributes a single force-closure score computed from the combined wrench space of its retained grasp candidates, rather than from point-wise completion variability or an individual grasp score alone.

In the present implementation, variability across these predictions is interpreted as an MC-dropout proxy for epistemic uncertainty in the completion model. The resulting decision rule is evaluated through controlled ablations against benchmark configurations that share the same overall perception, completion, grasp-generation, and execution pipeline, while selectively enabling or disabling individual components such as shape completion, geometric feasibility filtering, and uncertainty-aware abstention. This design keeps the experimental protocol and downstream system components fixed, allowing the contribution of each component to be assessed independently.

\section{Uncertainty-Aware Grasping Pipeline}
\label{methodology}

This paper presents an uncertainty-aware grasping framework designed for severely occluded agricultural environments. For a fixed partial point-cloud observation, the completion network is evaluated repeatedly with dropout active during inference, as it was during training. Variability across the resulting stochastic completion predictions is interpreted as an MC-dropout approximation of epistemic uncertainty in the learned completion model. We refer to this quantity as MC-dropout completion variability.

The framework adopts existing point-cloud completion and grasp-generation models. Its methodological contribution is a decision layer that propagates MC-dropout variability across completed shapes through grasp-candidate filtering and per-completion force-closure evaluation to an object-level attempt-or-abstain decision.

The current implementation does not explicitly parameterize aleatoric uncertainty arising from RGB-D measurement noise or the irreducible one-to-many ambiguity between a partial observation and its possible complete shapes. Consequently, epistemic and aleatoric uncertainty are not disentangled in the present implementation. The estimated MC-dropout variability is propagated to local grasp filtering and object-level attempt-or-abstain decision making, as indicated by the dashed connections in Fig.~\ref{fig:pipeline}. The complete algorithm is provided in the appendix in Algorithms~\ref{alg:strawberry_pipeline_1} and~\ref{alg:strawberry_pipeline_2}.

\begin{figure}
    \centering
    \includegraphics[width=1\linewidth]{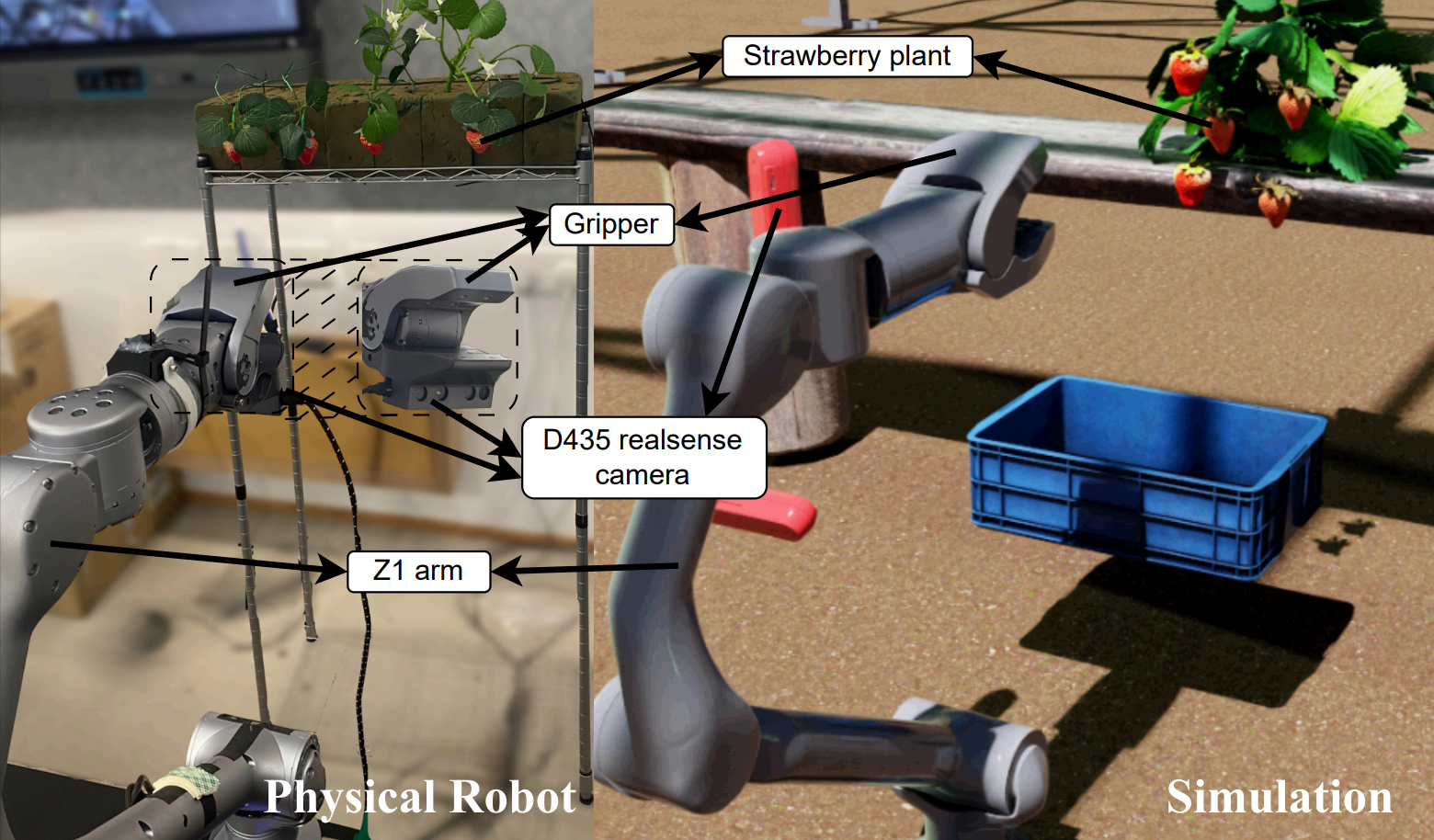}
    \caption{(Left) Physical robot setup replicating an indoor greenhouse strawberry plantation. (Right) Simulated strawberry field in NVIDIA Isaac Sim. The strawberry plant, Unitree Z1 robotic arm, grasping mechanism, and Intel RealSense D435i RGB-D camera are annotated.}
    \label{fig:sim_real_setup_labels}
\end{figure}


\subsection{Instance Point-Cloud Preprocessing}
\label{sec:preprocessing}
Given spatially aligned RGB and depth images, we apply a $5\times5$-pixel median filter to the depth image to suppress isolated depth noise. Valid depth pixels are then back-projected into 3D using the calibrated camera intrinsics. The resulting scene point cloud is voxel-downsampled to reduce noise and computational cost. Strawberry detections are refined into binary instance masks and used to extract a separate point cloud for each detected fruit. Invalid measurements and statistical outliers are subsequently removed. The resulting partial, denoised point cloud, $\mathcal{P}_{\mathrm{denoised}}$, is provided as input to the shape completion network. Full preprocessing definitions and parameter settings are provided in Appendix~\ref{app:preprocessing}.

\subsection{Point Cloud Completion and Epistemic Uncertainty Approximation}
\label{sec:completion_uncertainty}

Partial and occluded point clouds are completed using
PointAttN~\cite{PointAttn}, a transformer-based point-cloud completion network. Given a partial denoised observation $\mathcal{P}_{\mathrm{denoised}}$, the network predicts a completed point cloud $\mathcal{P}_{\mathrm{completed}}$.

Following Kendall and Gal~\cite{kendall_gal_uncertainty}, predictive uncertainty can be conceptually decomposed as

\begin{equation}
\begin{split}
\operatorname{Var}
\left[
Y \mid X,\mathcal{D}
\right]
={}&
\underbrace{
\mathbb{E}_{q(W)}
\left[
\operatorname{Var}
\left(
Y \mid X,W
\right)
\right]
}_{\text{aleatoric uncertainty}}
\\
&+
\underbrace{
\operatorname{Var}_{q(W)}
\left[
\mathbb{E}
\left(
Y \mid X,W
\right)
\right]
}_{\text{epistemic uncertainty}},
\end{split}
\label{eq:predictive_uncertainty_decomposition}
\end{equation}

where $X$ denotes the partial observation, $Y$ denotes the completed
geometry, $\mathcal{D}$ is the training set, $W$ denotes the completion-network parameters, and $q(W)$ approximates the posterior distribution $p(W \mid \mathcal{D})$ over the completion-network parameters.

In our implementation, the completion network does not predict an observation-noise variance or an explicit conditional distribution over all complete shapes compatible with the partial observation. Therefore, the aleatoric term in Eq.~\eqref{eq:predictive_uncertainty_decomposition} is not separately estimated. Instead, we approximate the epistemic term using MC dropout \cite{gal_gahramani_dropout}. Dropout layers used during training are retained at inference time. Dropout remains active during inference. For the same partial point cloud, we perform $K$ stochastic forward passes, with a newly and independently sampled dropout mask in each pass:

\begin{equation}
\begin{aligned}
\mathcal{P}_{\mathrm{completed}}^{(k)}
&=
f_{W^{(k)}}\!
\left(
\mathcal{P}_{\mathrm{denoised}}
\right),
\\
W^{(k)}
&\sim q(W),
\qquad k=1,\ldots,K.
\end{aligned}
\label{eq:mc_dropout_completions}
\end{equation}

Here, $f$ denotes the trained completion network, and $W^{(k)}$ denotes the effective network configuration induced by the dropout mask sampled during the $k$-th forward pass. Under the MC-dropout interpretation, $W^{(k)} \sim q(W)$ denotes the $k$-th sample from this approximate posterior. Because a new dropout mask is sampled for each forward pass, the $K$ stochastic passes produce $K$ effective network configurations and, consequently, $K$ completed point-cloud samples for the same partial observation. Collectively, these samples provide a Monte Carlo approximation of the predictive distribution induced by $q(W)$.
 
This produces the set of $K$ stochastic completion predictions

\begin{equation}
\left\{
\mathcal{P}_{\mathrm{completed}}^{(k)}
\right\}_{k=1}^{K}.
\end{equation}

Each stochastic completion contains $N$ reconstructed points,

\begin{equation}
\mathcal{P}_{\mathrm{completed}}^{(k)}
=
\left\{
p_n^{(k)}
\right\}_{n=1}^{N},
\qquad
p_n^{(k)} \in \mathbb{R}^{3}.
\label{eq:completion_points}
\end{equation}

For each output point index $n$, the mean predicted 3D location of output point n across the $K$ stochastic predictions is

\begin{equation}
\bar{p}_n
=
\frac{1}{K}
\sum_{k=1}^{K}
p_n^{(k)}.
\label{eq:mc_point_mean}
\end{equation}

We define the MC-dropout completion variability at point index $n$ as

\begin{equation}
u_n
=
\sqrt{
\frac{1}{K-1}
\sum_{k=1}^{K}
\left\|
p_n^{(k)}-\bar{p}_n
\right\|_2^2
}.
\label{eq:point_variability}
\end{equation}

The quantity $u_n$ measures spatial disagreement across stochastic completion predictions and is interpreted as a proxy for epistemic uncertainty in the completion model. A small $u_n$ means the model consistently places point n in nearly the same location. A large $u_n$ means its location changes considerably across dropout passes. These point-level variability values are used by the global and local filters defined below.

\paragraph{Global Uncertainty Filter}
The global uncertainty filter is defined as the mean point-level variability
over the completed strawberry:

\begin{equation}
U_{\mathrm{global}}
=
\frac{1}{N}
\sum_{n=1}^{N}
u_n.
\label{eq:global_variability}
\end{equation}

The strawberry continues to candidate-level filtering and grasp evaluation
only when

\begin{equation}
U_{\mathrm{global}}
\leq
\delta_{\mathrm{global}},
\label{eq:global_filter}
\end{equation}

where $U_{\mathrm{global}}$ is the global uncertainty of the completed strawberry and $\delta_{\mathrm{global}}$ is the maximum allowable global uncertainty. Completions with uncertainty above this threshold are rejected.

\begin{figure}
    \centering
    \includegraphics[width=\linewidth]{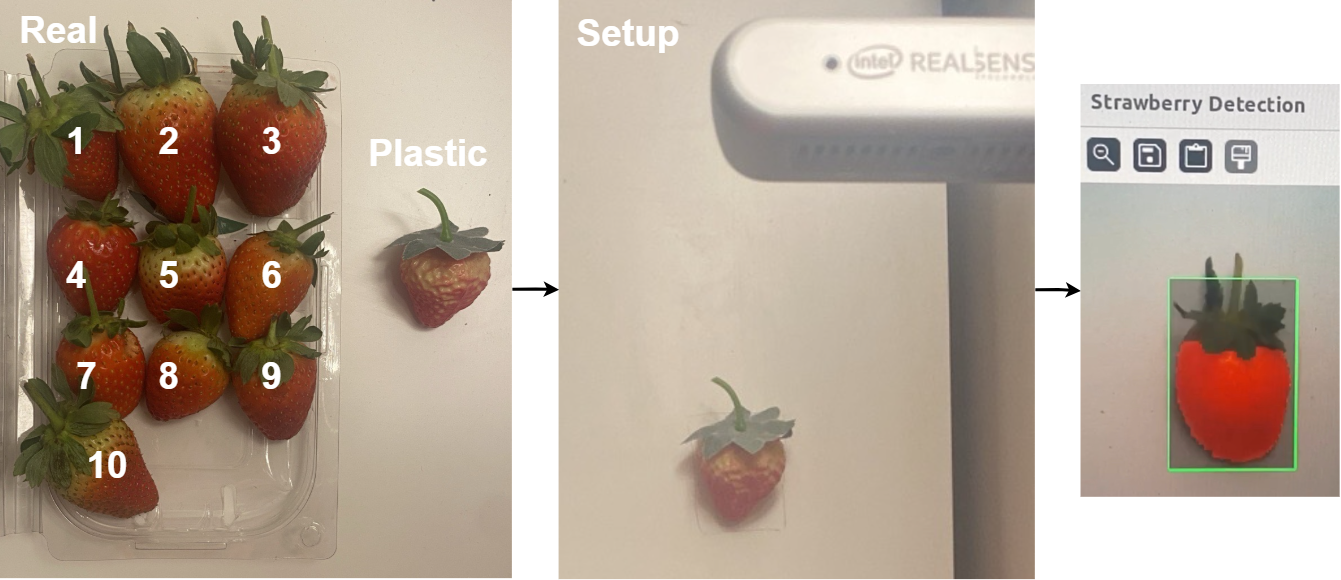}
    \caption{Experimental setup and strawberry detection pipeline. (Left) Top view of the ten real strawberries and the plastic strawberry that we measured the depth of. (Center) Depth measurement setup using the RealSense depth camera capturing strawberries placed on a fixed surface. (Right) Example output of the proposed detection system, showing a strawberry localized with a bounding box in the RGB image.}
    \label{fig:strawberry_setup}
\end{figure}

\subsection{Grasp Generation and Filtering}
\subsubsection{Grasp Generation}
For each completed point cloud $\mathcal{P}_{\text{completed}}^{(k)}$, we generate a set of candidate grasps using CGNet, which predicts 6-DoF grasp poses $\mathbf{G}_i^{(k)} \in \mathrm{SE}(3)$ along with per-grasp confidence scores $s_i^{(k)}$.

Formally, grasp generation is given by
\begin{equation}
\{\mathbf{G}_i^{(k)}, s_i^{(k)}\}_{i=1}^{M} =
\mathrm{CGNet}(\mathcal{P}_{\text{completed}}^{(k)}),
\label{eq:cgnet}
\end{equation}
where $\mathrm{CGNet}(\cdot)$ is the grasp generation network and $i$ indexes the individual grasp proposals from the number of grasp candidates $M$ generated for each completion sample $k$.

\begin{figure}
    \centering
    \includegraphics[width=\linewidth]{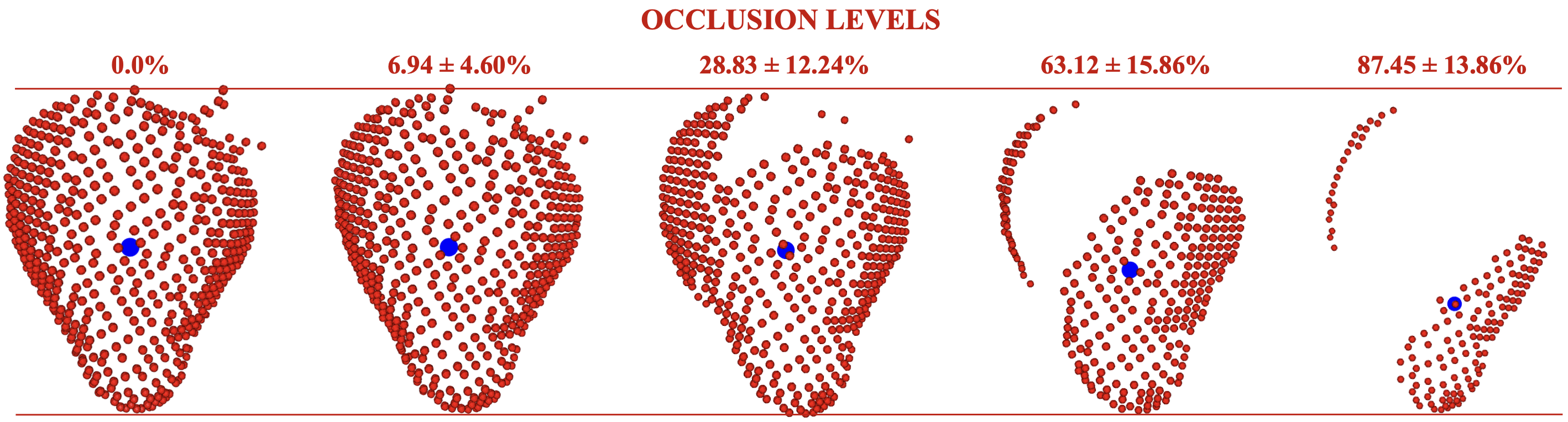}
    \caption{Representative partial strawberry point clouds under increasing synthetic leaf occlusion. The reported percentages denote the measured fraction of the fruit surface occluded at each level. Red points represent the visible strawberry geometry, and the blue point indicates the estimated fruit center.}
    \label{fig:qualitative_occlusions}
\end{figure}

\subsubsection{Filtering Pipeline}
\label{subsec:grasp_filtering}
Using the $K$ stochastic completion samples, we apply a multi-stage filtering procedure to the generated grasp candidates. The procedure comprises a local grasp uncertainty filter and two geometry-based filters, namely the grasp orientation (approach-direction and vertical-grasp) and jaw-fruit clearance filters.

\paragraph{Local Uncertainty Filter}
For strawberries that pass the global uncertainty filter, we apply a local uncertainty filter at the grasp level. For a candidate grasp $\mathbf{G}_i^{(k)}$, let
$\mathcal{R}(\mathbf{G}_i^{(k)})$ denote the gripper closing region. The
indices of the completion points lying within this region are

\begin{equation}
\mathcal{I}_i^{(k)}
=
\left\{
n
\;\middle|\;
p_n^{(k)}
\in
\mathcal{R}(\mathbf{G}_i^{(k)})
\right\}.
\label{eq:local_point_indices}
\end{equation}

Here, $\mathcal{I}_i^{(k)}$ includes the set of completed point indices $n$ such that the point $p_n^{(k)}$ lies inside the local region
$\mathcal{R}(\mathbf{G}_i^{(k)})$ associated with grasp candidate $\mathbf{G}_i^{(k)}$.

The local uncertainty or completion-variability score is

\begin{equation}
U_{\mathrm{local}}
\left(
\mathbf{G}_i^{(k)}
\right)
=
\frac{1}{
\left|
\mathcal{I}_i^{(k)}
\right|
}
\sum_{n\in\mathcal{I}_i^{(k)}}
u_n.
\label{eq:local_variability}
\end{equation}

A candidate grasp is retained only when

\begin{equation}
U_{\mathrm{local}}
\left(
\mathbf{G}_i^{(k)}
\right)
\leq
\delta_{\mathrm{local}},
\label{eq:local_filter}
\end{equation}

where $U_{\mathrm{local}}(\mathbf{G}_i^{(k)})$ is the uncertainty in the local point-cloud region associated with grasp candidate $i$ from completion sample $k$, and $\delta_{\mathrm{local}}$ is the maximum allowable local uncertainty. Candidates exceeding this threshold are discarded. This filter removes candidates whose gripper closing regions contain points with high disagreement across the stochastic completion predictions.

\paragraph{Approach Direction Filter}
To favor frontal access to the strawberry and reduce the risk of contact with the fruit or surrounding plant, grasps whose approach directions deviate excessively from the predefined desired approach direction are rejected. The filter is formulated as

\begin{equation}
\text{grasp passes}
\quad\Longleftrightarrow\quad
\mathbf{a}_i^{\top}\mathbf{f}
\geq
\theta_{\mathrm{dot}},
\end{equation}

where $\mathbf{a}_i=\mathbf{G}_i[0\!:\!3,2]$ is the approach direction, defined as the $z$-axis of the grasp frame, $\mathbf{f}$ is the predefined unit vector specifying the desired frontal approach direction, and $\theta_{\mathrm{dot}}$ is the minimum allowable cosine similarity. Both vectors are expressed in the same coordinate frame and normalized to unit length.

\paragraph{Vertical Grasp Filter}
Grasps whose jaw-opening axes are strongly aligned with the world vertical direction are rejected according to

\begin{equation}
\text{grasp rejected}
\quad\Longleftrightarrow\quad
\left|
\mathbf{G}_i[0\!:\!3,0]^{\top}\mathbf{z}
\right|
>
\theta_{\mathrm{vert}},
\end{equation}

where $\mathbf{G}_i[0\!:\!3,0]$ is the grasp-frame $x$-axis, corresponding to the jaw-opening direction, $\mathbf{z}$ is the unit world-vertical vector, and $\theta_{\mathrm{vert}}$ is the maximum allowable absolute cosine similarity with the vertical direction. The grasp pose and vertical vector are expressed in the same coordinate frame. The absolute value ensures that both upward and downward vertical alignments are treated identically. This task-specific constraint removes vertically stacked finger configurations that were found to be less reliable in our setup.

\paragraph{Jaw-Object Intersection Filter}
For each grasp, we verify that the gripper jaws do not intersect the object through a geometric clearance check to ensure that the gripper jaws do not damage the strawberries as illustrated in \textit{Filtering d)} in Fig. \ref{fig:pipeline}.

The jaw lines are
\begin{align}
\mathbf{L}_{\text{left}} &: \mathbf{c} - \frac{w}{2}\mathbf{x} + t\mathbf{a}, \\
\mathbf{L}_{\text{right}} &: \mathbf{c} + \frac{w}{2}\mathbf{x} + t\mathbf{a},
\end{align}
where $\mathbf{c} = \mathbf{G}_i[0:3, 3]$ is the grasp center, $\mathbf{x} = \mathbf{G}_i[0:3, 0]$ is the jaw-opening direction, $w$ is the gripper width, $\mathbf{a}$ is the approach direction, and $t$ is the jaw length.

A grasp is considered valid if no point in $\mathcal{P}_{\text{completed}}$ lies within a distance $\tau$\ of either jaw line as formulated in 
\begin{equation}
\text{grasp passes} \iff \forall \mathbf{p} \in \mathcal{P}: \min(d(\mathbf{p}, \mathbf{L}_{\text{left}}), d(\mathbf{p}, \mathbf{L}_{\text{right}})) > \tau .
\label{eq:jaw_clearance}
\end{equation}
In this equation, $\mathbf{p} \in \mathcal{P}$ denotes a point in the completed point cloud, $\mathbf{L}_{\text{left}}$ and $\mathbf{L}_{\text{right}}$ are the left and right jaw lines, $d(\mathbf{p}, \mathbf{L})$ is the shortest Euclidean distance from point $\mathbf{p}$ to line $\mathbf{L}$, and $\tau$ is the minimum allowable clearance threshold.

This tolerance serves as a geometric clearance margin during collision checks against reconstructed point clouds, reflecting uncertainty in reconstruction accuracy and preventing gripper jaws from being placed too close to potentially inaccurate surface estimates.

Together, these filters remove unreliable grasp candidates and support object-level abstention under high uncertainty, which is central to the proposed decision framework. After filtering, only $M' \leq M$ of the original $M$ grasp proposals remain for each completion sample,
\begin{equation}
\{\mathbf{G}_j^{(k)}, s_j^{(k)}\}_{j=1}^{M'_k} \subseteq \{\mathbf{G}_i^{(k)}, s_i^{(k)}\}_{i=1}^{M},
\end{equation}
where $\mathbf{G}_i^{(k)}$ and $s_i^{(k)}$ denote the original $M$ grasp proposals and scores for completion $k$, $\mathbf{G}_j^{(k)}$ and $s_j^{(k)}$ denote the remaining grasps after filtering, and $M'_k \le M$ is the number of grasps that remain for completion $k$. The remaining question is how to conservatively assess grasp feasibility across uncertain shape completions, which we address using physically grounded force-closure grasp quality metrics.

\begin{equation}
\{\{\mathbf{G}_j^{(k)}, s_j^{(k)}\}_{j=1}^{M'_k}\}_{k=1}^{K} \subseteq \{\{\mathbf{G}_i^{(k)}, s_i^{(k)}\}_{i=1}^{M}\}_{k=1}^{K} ,
\end{equation}

\begin{table}
\centering
\caption{Each row of the ablation methods indicates which components of the grasping pipeline are enabled.}
\label{tab:ablation_config}
\footnotesize
\setlength{\tabcolsep}{5pt}
\renewcommand{\arraystretch}{1.15}

\resizebox{\columnwidth}{!}{%
\begin{tabular}{lcccc}
\hline
\textbf{Method} &
\makecell{\textbf{Shape} \\ \textbf{Completion}} &
\makecell{\textbf{CGNet}} &
\makecell{\textbf{Geometric} \\ \textbf{Filtering}} &
\makecell{\textbf{Uncertainty} \\ \textbf{Filtering}} \\
\hline

CGNet (Partial)
& \xmark & \cmark & \xmark & \xmark \\

CGNet+Geom (Partial)
& \xmark & \cmark & \cmark & \xmark \\

Centroid (Completed)
& \cmark & \xmark & \xmark & \xmark \\ \hline

CGNet (Completed)
& \cmark & \cmark & \xmark & \xmark \\

CGNet+Geom (Completed)
& \cmark & \cmark & \cmark & \xmark \\

UNCLE-Grasp \textit{(Ours)}
& \cmark & \cmark & \cmark & \cmark \\
\hline
\end{tabular}
}
\end{table}

\subsection{Grasp Quality Metrics}
\label{LCB}
For each completion sample $k$, CGNet may produce multiple grasp candidates with associated confidence scores. Since these scores are learned and can be unreliable under severe occlusion, we do not select grasps based on the CGNet score alone. Instead, we evaluate the physical feasibility of each remaining grasp using the force-closure $\epsilon$ metric, which characterizes a grasp’s ability to resist arbitrary external forces and torques. Each contact contributes a friction cone parameterized by a friction coefficient $\mu$, and the resulting contact wrenches form a convex set in wrench space. We use the $\epsilon$ metric, defined as the radius of the largest ball centered at the origin contained within this convex hull, with larger values indicating greater grasp robustness.

\subsubsection{Contact Estimation}
For each remaining grasp, we estimate the contact points by stepping along the approach direction until the jaw tips reach the point cloud surface. Given a grasp pose $\mathbf{G}$ and jaw position trajectories, we compute contact points $\{\mathbf{c}_{\text{left}}, \mathbf{c}_{\text{right}}\}$ and their associated surface normals $\{\mathbf{n}_{\text{left}}, \mathbf{n}_{\text{right}}\}$ using nearest-neighbor queries.

\subsubsection{Grasp Epsilon Metric}
For each contact pair, we compute the grasp quality using the epsilon metric (force closure measure). The friction cone is discretized into $N_{dir}$ directions, providing a conservative approximation that avoids overestimating grasp stability, particularly under uncertain contact conditions.

A contact wrench represents the force and torque applied to the object at a contact point by the gripper. It is constructed as
\begin{equation}
\mathbf{W} = \begin{bmatrix} \mathbf{F}_1 & \cdots & \mathbf{F}_{2N_{dir}} \\ \boldsymbol{\tau}_1 & \cdots & \boldsymbol{\tau}_{2N_{dir}} \end{bmatrix} \in \mathbb{R}^{6 \times 2N_{dir}}.
\end{equation}
Each column of $\mathbf{W}$ corresponds to one contact wrench, consisting of a contact force $\mathbf{F}_j \in \mathbb{R}^3$ and its corresponding torque $\boldsymbol{\tau}_j = \mathbf{c}_j \times \mathbf{F}_j \in \mathbb{R}^3$, where $\mathbf{c}_j$ is the contact point, $N_{dir}$ is the number of discretized friction-cone directions per contact, and $2N_{dir}$ accounts for both contacts from the gripper on the object.

Epsilon quantifies the robustness of a grasp to external forces and torques. It is defined as the minimum distance from the zero wrench, representing no applied force or torque, to the boundary of the grasp wrench space. Equivalently, it is the radius of the largest origin-centered ball contained within that space and represents the smallest normalized disturbance magnitude, across all wrench directions, that brings the grasp to its stability limit. A larger epsilon therefore indicates greater static grasp robustness. This value is computed as
\begin{equation}
\epsilon = \min_{j} \frac{|d_j|}{\|\mathbf{n}_j\|_2},
\end{equation}
where the set of feasible contact wrenches is represented as the intersection of halfspaces
$\mathbf{n}_j^\top \mathbf{w} + d_j \le 0$.
Here, $\mathbf{n}_j$ is the outward normal vector of the $j$-th supporting hyperplane in wrench space, $d_j$ is its offset from the origin, $\|\mathbf{n}_j\|_2$ is the Euclidean norm of $\mathbf{n}_j$, and $\epsilon$ is the resulting force-closure robustness margin.

\begin{table*}
\centering
\caption{Overall grasp success rate: Successful grasps divided by the total number of strawberries in the scene under varying occlusion levels. Reported values are mean $\pm$ standard deviation.}
\label{tab:success_over_total}

\resizebox{\textwidth}{!}{%

\begin{tabular}{lccccc|c}
\hline
& \multicolumn{6}{c}{Occlusion Percentage} \\
\cline{2-7}
Method
& 0\%
& $6.94 \pm 4.60\%$
& $28.83 \pm 12.24\%$
& $63.12 \pm 15.86\%$
& $87.45 \pm 13.86\%$
& Real Leaf ($\sim70\%$) \\
\hline
\hline
\multicolumn{7}{l}{\textit{\textbf{Simulation}}} \\
\hline

CGNet (Partial)
& $0.240 \pm 0.080$
& $0.280 \pm 0.098$
& $0.280 \pm 0.133$
& $0.300 \pm 0.100$
& $0.240 \pm 0.080$
& -- \\

CGNet+Geom (Partial)
& $0.440 \pm 0.080$
& $0.440 \pm 0.080$
& $0.440 \pm 0.080$
& $0.460 \pm 0.128$
& $0.440 \pm 0.080$
& -- \\

Centroid (Completed)
& $0.440 \pm 0.080$
& $0.400 \pm 0.126$
& $0.560 \pm 0.080$
& $0.600 \pm 0.155$
& $0.540 \pm 0.156$
& -- \\

\hline

CGNet (Completed)
& $0.560 \pm 0.196$
& $0.500 \pm 0.205$
& $0.540 \pm 0.180$
& $0.600 \pm 0.179$
& $0.680 \pm 0.098$
& -- \\

CGNet+Geom (Completed)
& \textbf{0.600 $\pm$ 0.179}
& \textbf{0.660 $\pm$ 0.156}
& \textbf{0.620 $\pm$ 0.166}
& \textbf{0.660 $\pm$ 0.180}
& $0.680 \pm 0.098$
& -- \\

UNCLE-Grasp \textit{(Ours)}
& $0.360 \pm 0.150$
& $0.280 \pm 0.098$
& $0.200 \pm 0.000$
& $0.500 \pm 0.134$
& \textbf{0.740 $\pm$ 0.092}
& -- \\

\hline
\hline
\multicolumn{7}{l}{\textit{\textbf{Physical Robot}}} \\
\hline

CGNet (Completed)
& \textbf{0.450 $\pm$ 0.292}
& $0.400 \pm 0.122$
& \textbf{0.550 $\pm$ 0.332}
& \textbf{0.500 $\pm$ 0.158}
& \textbf{0.550 $\pm$ 0.187}
& $0.100 \pm 0.122$ \\

CGNet+Geom (Completed)
& \textbf{0.450 $\pm$ 0.187}
& \textbf{0.550 $\pm$ 0.292}
& $0.400 \pm 0.300$
& $0.250 \pm 0.224$
& $0.400 \pm 0.122$
& \textbf{0.200 $\pm$ 0.187} \\

UNCLE-Grasp \textit{(Ours)}
& $0.250 \pm 0.158$
& $0.350 \pm 0.122$
& $0.450 \pm 0.292$
& \textbf{0.500 $\pm$ 0.224}
& $0.200 \pm 0.100$
& \textbf{0.200 $\pm$ 0.100} \\

\hline
\end{tabular}%
}
\end{table*}

\paragraph{Aggregation Across Uncertain Completions}

For each completion sample $k$, the wrench contributions of all grasp candidates that remain after filtering are combined to construct a single grasp wrench space. The force-closure metric $\epsilon_k$ is then computed from this combined space, yielding one object-level feasibility score for each completed shape. If no grasp survives filtering, we set $\epsilon_k = 0$. A larger $\epsilon_k$ indicates that the retained grasp set provides broader force-torque coverage, whereas $\epsilon_k = 0$ indicates that the completion does not yield a feasible retained grasp set under the adopted evaluation. In this way, each completion contributes one object-level feasibility score indicating whether the reconstructed object admits at least one stable grasp.

Letting $\mathcal{G}_k = \{\mathbf{G}_j^{(k)}, s_j^{(k)}\}_{j=1}^{M'_k}$ denote the set of grasps for completion $k$ that remain after filtering, we compute one force-closure value for the combined grasp set as
\begin{equation}
\epsilon_k
=
\begin{cases}
\epsilon\!\left(\mathcal{G}_k\right),
& \mathcal{G}_k \neq \varnothing,\\
0,
& \mathcal{G}_k = \varnothing.
\end{cases}
\label{eq:epsilon_per_completion}
\end{equation}

$\epsilon_k$ is the maximum force-closure value for that completion (or zero if no grasp survives).

From $\{\epsilon_k\}_{k=1}^{K}$ we compute
\begin{align}
\bar{\epsilon} &= \frac{1}{K}\sum_{k=1}^{K} \epsilon_k, \\
\sigma_{\epsilon} &= \sqrt{\frac{1}{K-1}\sum_{k=1}^{K}(\epsilon_k - \bar{\epsilon})^2}.
\end{align}
Here, $\epsilon_k$ is the force-closure score for completion $k$, $K$ is the total number of stochastic completions, $\bar{\epsilon}$ is the sample mean force-closure score across completions, and $\sigma_{\epsilon}$ is their corresponding sample standard deviation.

We then define the lower confidence bound on grasp feasibility as the LCB metric:
\begin{equation}
\mathrm{LCB} = \bar{\epsilon} - z_{\alpha}\sigma_{\epsilon}.
\label{eq:lcb}
\end{equation}
We do not rely on $\sigma_{\epsilon}$ alone because a small standard deviation indicates only that the completion samples produce similar feasibility scores; those scores may still be consistently close to zero. The LCB therefore combines the average grasp feasibility with its consistency across uncertain completions, rewarding large values of $\bar{\epsilon}$ while penalizing disagreement through $z_{\alpha}\sigma_{\epsilon}$. A grasp is attempted only if $\mathrm{LCB}>0$, indicating that the retained grasp sets maintain positive force-closure feasibility after accounting for completion-induced variability; otherwise, the system abstains. We use a one-sided lower bound because the decision depends on whether the force-closure margin remains positive after accounting for variation across completion samples, hence larger positive margins are more desirable since they indicate a more robust retained grasp set.

The confidence scaling factor $z_{\alpha}$ is a linear function of the estimated leaf occlusion level $\alpha$, bounded by the minimum, $z_{\min}$, and maximum, $z_{\max}$, confidence scaling parameters corresponding to the least and most occluded cases respectively, and $\alpha_{\max}$ is the maximum possible occlusion level. In particular, we use a linear scaling,
\begin{equation}
z_{\alpha} = z_{\min} + \frac{\alpha}{\alpha_{\max}}\left(z_{\max} - z_{\min}\right),
\end{equation}
which relaxes the bound in easier (low-occlusion) cases and becomes stricter as occlusion increases.

\section{Experiments}
We evaluate UNCLE-Grasp through controlled component ablations in simulation and physical-robot experiments under leaf occlusion. The experiments assess the contributions of shape completion, geometric filtering, MC-dropout completion variability, and object-level abstention within a common implementation and evaluation protocol. We first describe the experimental setup and implementation details, followed by the occlusion-generation procedure, trial protocol, and evaluation metrics.

\subsection{Experimental Setup}
\label{sec:experimental_setup}

\subsubsection{Hardware and Simulation Environments}
The system interfaces with an Intel RealSense D435i RGB-D camera mounted on a Unitree Z1 robotic arm, as shown in Fig.~\ref{fig:sim_real_setup_labels}. Physical robot experiments were conducted in a controlled indoor environment arranged to replicate an indoor greenhouse strawberry plantation, with multiple hanging strawberries and varied occlusion patterns. 

All physical robot experiments were conducted using offboard control from the same workstation used for simulation experiments (Intel x86\_64 CPU with an NVIDIA RTX 5080 GPU). Simulation experiments were performed in NVIDIA Isaac Sim, replicating the physical robot setup, including the strawberry plant geometry, robot arm, gripper, and camera configuration. Identical robot hardware parameters and environmental conditions were used across simulation and physical experiments to ensure fair comparison across methods.

\begin{table*}
\centering
\caption{Attempt rate: Attempted strawberries divided by the total number of strawberries in the scene under varying occlusion levels. Attempted strawberries are defined as the number of detected \& filtered strawberries. Reported values are mean $\pm$ standard deviation.}
\label{tab:attempt_results}
\resizebox{\textwidth}{!}{%
\begin{tabular}{lccccc|c}
\hline
& \multicolumn{6}{c}{Occlusion Percentage} \\
\cline{2-7}
Method
& 0\%
& $6.94 \pm 4.60\%$
& $28.83 \pm 12.24\%$
& $63.12 \pm 15.86\%$
& $87.45 \pm 13.86\%$
& Real Leaf ($\sim70\%$) \\
\hline
\hline
\multicolumn{7}{l}{\textit{\textbf{Simulation}}} \\
\hline

CGNet (Partial)
& $0.700 \pm 0.100$
& $0.800 \pm 0.089$
& $0.700 \pm 0.100$
& $0.780 \pm 0.140$
& $0.720 \pm 0.133$
& -- \\

CGNet+Geom (Partial)
& $0.640 \pm 0.080$
& $0.680 \pm 0.133$
& $0.620 \pm 0.060$
& $0.640 \pm 0.080$
& $0.640 \pm 0.080$
& -- \\

Centroid (Completed)
& $0.900 \pm 0.100$
& \textbf{0.900 $\pm$ 0.134}
& \textbf{0.900 $\pm$ 0.100}
& \textbf{0.920 $\pm$ 0.098}
& \textbf{0.900 $\pm$ 0.100}
& -- \\
\hline

CGNet (Completed)
& \textbf{0.980 $\pm$ 0.060}
& $0.880 \pm 0.098$
& $0.840 \pm 0.150$
& $0.900 \pm 0.134$
& \textbf{0.900 $\pm$ 0.100}
& -- \\

CGNet+Geom (Completed)
& $0.900 \pm 0.100$
& $0.860 \pm 0.156$
& $0.880 \pm 0.098$
& \textbf{0.920 $\pm$ 0.098}
& $0.880 \pm 0.098$
& -- \\

UNCLE-Grasp \textit{(Ours)}
& $0.520 \pm 0.183$
& $0.280 \pm 0.098$
& $0.300 \pm 0.100$
& $0.600 \pm 0.126$
& $0.860 \pm 0.092$
& -- \\
\hline
\hline

\multicolumn{7}{l}{\textit{\textbf{Physical Robot}}} \\
\hline

CGNet (Completed)
& $0.900 \pm 0.122$
& \textbf{0.650 $\pm$ 0.200}
& \textbf{0.750 $\pm$ 0.274}
& \textbf{0.750 $\pm$ 0.224}
& \textbf{0.950 $\pm$ 0.100}
& $0.300 \pm 0.100$ \\

CGNet+Geom (Completed)
& \textbf{0.950 $\pm$ 0.100}
& \textbf{0.650 $\pm$ 0.200}
& $0.550 \pm 0.187$
& $0.500 \pm 0.224$
& $0.850 \pm 0.122$
& $0.450 \pm 0.187$ \\

UNCLE-Grasp \textit{(Ours)}
& $0.300 \pm 0.100$
& $0.350 \pm 0.122$
& $0.450 \pm 0.292$
& $0.550 \pm 0.292$
& $0.250 \pm 0.000$
& \textbf{0.600 $\pm$ 0.300} \\
\hline

\end{tabular}%
}
\end{table*}

\subsubsection{Experimental Targets and Depth Validation}
For safety, repeatability, and cost considerations, all grasping experiments were conducted using high-fidelity plastic strawberries. To ensure that the perception and depth-based components of the pipeline remain valid, we conducted additional measurements comparing the depth values obtained from a plastic and 10 real strawberries at equivalent distances as shown in Fig.~\ref{fig:strawberry_setup}.

While the visual and geometric properties of these strawberries closely resemble real fruit, their mechanical properties differ. In particular, failed grasps manifest differently: in real strawberries, excessive gripping force would lead to visible deformation or squishing, whereas in plastic strawberries, failure results in slippage and subsequent falling. From a manipulation evaluation perspective, both cases correspond to unsuccessful grasps. Therefore, the grasp success metric remains equivalent regardless of whether real or artificial strawberries are used.

\subsubsection{Implementation Details}

Raw RGB-D point clouds are cleaned by removing invalid values (NaNs and infinities) and statistical outliers beyond three standard deviations from the median. Strawberry instances are detected using a two-stage pipeline trained on diverse strawberry images, achieving robust localization under varying lighting conditions and partial occlusions: YOLOv8~\cite{yolov8_ultralytics} provides coarse 2D bounding boxes, which are refined using SAM2~\cite{sam2} to obtain precise instance masks. Each mask is projected into 3D to extract a segmented partial point cloud $\mathcal{P}$ corresponding to an individual strawberry.

\subsection{Evaluation Protocol}
\label{sec:evaluation_protocol}

\subsubsection{Occlusion Levels and Trial Protocol}
\label{Occlusion Levels and Trial Protocol}
To systematically evaluate grasping performance under partial observability, we apply the same synthetic leaf occlusion procedure to strawberry point clouds in both simulation and physical robot experiments. Using the geometric leaf model described in Appendix~\ref{leaf_model}, points are artificially removed from both simulated and real RGB-D strawberry point clouds to produce controlled and reproducible occlusion levels that resemble real leaf coverage. This controlled occlusion enables repeatable evaluation that would be difficult to achieve using physical leaves alone.

Experiments are conducted across five synthetically generated occlusion levels corresponding to the following empirical point removal rates: $0\%,$ 
$6.94 \pm 4.60\%,$
$28.83 \pm 12.24\%,$ 
$63.12 \pm 15.86\%,$ and 
$87.45 \pm 13.86\%.$
Qualitatively, the partial point clouds
at these occlusion levels are shown in
Fig.~\ref{fig:qualitative_occlusions}, illustrating the progressive
loss of visible strawberry geometry as occlusion increases.

In addition to these controlled settings, we evaluate performance under real leaf occlusion on the physical robot with natural leaves covering approximately 70\% of each strawberry, as shown in the physical robot setup in Fig.~\ref{fig:sim_real_setup_labels}. This setting is reported separately in Table~\ref{tab:occlusion_results} as \emph{Real Leaf ($\sim$70\%)} and is intended to assess realism rather than exact correspondence with synthetic occlusion levels.

For each occlusion level, we perform ten trials in simulation, each consisting of grasp attempts on five strawberries, and five trials on the physical robot, each with grasp attempts on four strawberries. Across all grasp selection strategies, this results in a total of 1500 simulated grasp attempts and 360 physical robot grasp attempts.

\subsection{Grasp Selection Strategies}

As Table~\ref{tab:ablation_config} summarizes, we evaluate the effectiveness of uncertainty-aware grasp selection under partial occlusion through a controlled comparison of six grasp selection strategies designed to isolate the causal contributions of individual pipeline components: CGNet (Partial), CGNet+Geom (Partial), Centroid (Completed), CGNet (Completed), CGNet+Geom (Completed), and UNCLE-Grasp. 

The first three ablations disentangle the roles of shape completion, grasp generation, and geometric constraints under increasing occlusion. They are evaluated in simulation only to enable controlled, repeatable comparison of intentionally simplified pipeline variants without incurring the time and hardware cost of physical execution.

\begin{table*} 
\centering
\caption{Conditional grasp success rate: Successful grasps divided by the number of attempted strawberries under varying occlusion levels. Reported values are mean $\pm$ standard deviation.}
\label{tab:occlusion_results}
\resizebox{\textwidth}{!}{%
\begin{tabular}{lccccc|c}
\hline
& \multicolumn{6}{c}{Occlusion Percentage} \\
\cline{2-7}
Method
& 0\%
& $6.94 \pm 4.60\%$
& $28.83 \pm 12.24\%$
& $63.12 \pm 15.86\%$
& $87.45 \pm 13.86\%$
& Real Leaf ($\sim70\%$) \\
\hline 
\hline
\multicolumn{7}{l}{\textit{\textbf{Simulation}}} \\
\hline 
CGNet (Partial)
& $0.342 \pm 0.087$
& $0.348 \pm 0.110$
& $0.417 \pm 0.233$
& $0.380 \pm 0.092$
& $0.337 \pm 0.093$ 
& -- \\

CGNet+Geom (Partial)
& $0.683 \pm 0.033$
& $0.652 \pm 0.061$
& $0.708 \pm 0.100$
& $0.708 \pm 0.100$
& $0.683 \pm 0.033$ 
& -- \\

Centroid (Completed)
& $0.495 \pm 0.106$
& $0.443 \pm 0.134$
& $0.625 \pm 0.090$
& $0.655 \pm 0.157$
& $0.600 \pm 0.167$ 
& -- \\
\hline

CGNet (Completed)
& $0.580 \pm 0.227$
& $0.585 \pm 0.263$
& $0.647 \pm 0.197$
& $0.690 \pm 0.247$
& $0.760 \pm 0.107$
& -- \\

CGNet+Geom (Completed)
& $0.675 \pm 0.216$
& $0.777 \pm 0.174$
& $0.710 \pm 0.196$
& $0.730 \pm 0.228$
& $0.780 \pm 0.129$
& -- \\

UNCLE-Grasp \textit{(Ours)}
& \textbf{0.742 $\pm$ 0.280}
& \textbf{1.000 $\pm$ 0.000}
& \textbf{0.750 $\pm$ 0.250}
& \textbf{0.850 $\pm$ 0.213}
& \textbf{0.870 $\pm$ 0.140}
& -- \\
\hline
\hline
\multicolumn{7}{l}{\textit{\textbf{Physical Robot}}} \\
\hline 
CGNet (Completed)
& $0.500 \pm 0.293$
& $0.683 \pm 0.291$
& $0.617 \pm 0.332$
& $0.683 \pm 0.186$
& $0.600 \pm 0.255$
& $0.300 \pm 0.400$ \\

CGNet+Geom (Completed)
& $0.500 \pm 0.274$
& $0.733 \pm 0.389$
& $0.600 \pm 0.374$
& $0.467 \pm 0.400$
& $0.483 \pm 0.170$
& $0.367 \pm 0.371$ \\

UNCLE-Grasp \textit{(Ours)}
& \textbf{0.800 $\pm$ 0.400}
& \textbf{1.000 $\pm$ 0.000}
& \textbf{1.000 $\pm$ 0.000}
& \textbf{0.950 $\pm$ 0.100}
& \textbf{0.800 $\pm$ 0.400}
& \textbf{0.517 $\pm$ 0.410} \\
\hline
\end{tabular}%
}
\end{table*}

\subsubsection{CGNet (Partial)} 
We apply CGNet directly to generate and execute the top scoring grasp on partial point clouds without shape completion.

\subsubsection{CGNet+Geom (Partial)}
We apply geometric filtering to the grasps generated by CGNet and execute the top scoring one from the remaining grasps on the partial point cloud without shape completion.

\subsubsection{Centroid (Completed)}
We implement centroid-based grasping (instead of CGNet) on completed point clouds.

\subsubsection{CGNet (Completed)}
We apply CGNet to generate and execute the top scoring grasp on a completed point cloud. No uncertainty modeling, geometric filtering, or abstention mechanism is applied, resulting in a fast but deterministic grasp selection policy.

\subsubsection{CGNet+Geom (Completed)}
We apply geometric filtering to the grasps generated by CGNet and execute the top scoring one from the remaining grasps on the completed point cloud. We disable dropout during point cloud completion, and therefore do not produce uncertainty estimates. This ablation isolates the effect of geometric filtering in the absence of uncertainty-aware reasoning.

\subsubsection{UNCLE-Grasp}
This strategy evaluates grasp feasibility across multiple MC-dropout completion samples and applies the geometric filters used in the CGNet+Geom (Completed) strategy, uncertainty-aware filtering (global and local uncertainty filters), and the LCB decision rule described in Subsection~\ref{LCB}.

\paragraph{Practical Considerations for Real-Time Deployment}
Repeated point-cloud completion and grasp evaluation across the $K$ MC-dropout samples constitute the main computational bottleneck of UNCLE-Grasp. To maintain practical execution rates on the physical robot, the stochastic completion passes are parallelized using thread pools, nearest-neighbor queries during contact estimation are accelerated using KD-trees, and the friction-cone discretization $N_{\mathrm{dir\text{-}opt}}$ is reduced to lower the cost of wrench computation.

Although jaw-object intersection checking contributes only a relatively small portion of the overall runtime, it is further accelerated using a hierarchical strategy. Axis-aligned bounding boxes with padding $\tau$ are first constructed around each jaw line to reject distant points. Point-to-line distance checks are then restricted to points within these bounding boxes, and the remaining calculations are vectorized using NumPy broadcasting to avoid per-point Python loops.

These optimizations affect runtime only and do not alter the uncertainty estimation, filtering criteria, or object-level decision rule. Because simulation execution time is less constrained, these system-level optimizations are primarily applied during physical-robot deployment.

\subsection{Evaluation Metrics}

We report three complementary metrics. First, the overall grasp success rate is defined as the number of successful grasps divided by the total number of strawberries in the scene under varying occlusion levels. This metric directly measures the fraction of available strawberries that are successfully harvested. 

Second, the attempt rate is defined as the number of attempted strawberries divided by the total number of strawberries in the scene. An attempt corresponds to a detected strawberry that passes all filtering stages and for which the robot attempts to grasp. Strawberries missed during detection or segmentation, as well as those rejected by the filtering or abstention mechanisms, are not counted as attempts. Consequently, even baselines without additional filtering may have an attempt rate below 1 if some strawberries are not successfully detected. This metric therefore measures how frequently the method chooses to act on the available targets. 

Third, the conditional grasp success rate is defined as the number of successful grasps divided by the number of attempted strawberries (overall grasp success rate / attempt rate). A grasp is considered successful if the robot securely lifts the strawberry without slippage or collision. This metric measures grasp reliability conditioned on the system deciding to execute a grasp.

Together, the three metrics distinguish overall harvesting performance, selectivity, and conditional execution reliability. Overall success rate measures the proportion of all strawberries that are successfully harvested, attempt rate indicates how often the method chooses to act, and success over attempts quantifies the reliability of the selected actions. This joint evaluation is particularly important for UNCLE-Grasp because abstention is an intentional risk-aware behavior: success over attempts captures the benefit of avoiding geometrically uncertain grasps, while attempt rate and overall success rate characterize the corresponding coverage and total harvesting performance.

\definecolor{truepurple}{RGB}{120, 81, 140} 
\begin{figure} 
\centering
\begin{tikzpicture}

\begin{axis}[
    width=\linewidth,
    height=5.4cm,
    ylabel={MC-Dropout Completion Variability},
    xlabel={Occlusion level (\%)},
    xtick={0,1,2,3,4},
    xticklabels={0, 6.94, 28.83, 63.12, 87.45},
    ymin=0,
    grid=both,
    ymajorgrids=true,
    xmajorgrids=false,
    bar width=16pt,
    tick label style={font=\small},
    label style={font=\small}
]

\addplot+[
    ybar,
    fill=truepurple!85,
    draw=black,
    mark=none, 
    error bars/.cd,
    y dir=both,
    y explicit,
    error bar style={line width=1.0pt, black},
    error mark options={
        rotate=90,
        mark size=3pt,
        line width=1.0pt,
        black
    }
]
coordinates {
    (0,0.001176) +- (0,0.000029)
    (1,0.001304) +- (0,0.000022)
    (2,0.001308) +- (0,0.000045)
    (3,0.001502) +- (0,0.000162)
    (4,0.002110) +- (0,0.001361)
};

\end{axis}
\end{tikzpicture}
\caption{Completion uncertainty under increasing occlusion. Bars indicate mean uncertainty across MC-dropout samples and whiskers denote standard deviation. Higher values indicate greater disagreement among completion-model samples.}
\label{fig:completion_uncertainty_bar}
\end{figure}

\subsection{Hyperparameters and Thresholds}

Unless otherwise stated, all hyperparameters are shared across methods and fixed for all experiments.

\paragraph{Completion and Grasp Generation}
We use an MC dropout rate of $0.1$ with $K=20$ completion samples per observation. For each completed shape, $M=200$ candidate grasps are generated using CGNet.

\paragraph{Geometric Constraints}
The grasp orientation thresholds were selected empirically by testing a range of values and balancing grasp safety against allowing grasp variability. The approach-direction threshold $\theta_{\text{dot}} = 0.7$ restricts grasps to within approximately $45^\circ$ of the camera-facing direction ($\cos 45^\circ \approx 0.707$). Stricter limits rejected many feasible grasps, whereas wider limits admitted side and rear approaches that could collide with occluded leaves, stems, or neighboring strawberries. Similarly, the vertical-alignment threshold $\theta_{\text{vert}} = 0.5$ rejects grasps whose jaw-opening axis lies within $60^\circ$ of the vertical axis $\mathbf{z} = [0,0,1]^T$. This threshold excludes near-vertical finger configurations that are more prone to fruit slippage or damage while preserving a sufficiently wide range of valid non-vertical grasp orientations.

\paragraph{Gripper and Contact Parameters}
Based on the gripper specifications, the jaw width is set to $w = 0.04\,\text{m}$, and the jaw length is constrained to $t \in [0, 0.2]\,\text{m}$. A moderate friction coefficient $\mu = 0.5$ is used to reflect stable yet gentle grasps appropriate for delicate objects such as strawberries.

To account for depth sensing uncertainty, we use a conservative clearance margin of $\tau = 5\,\text{mm}$. Prior work shows that RGB-D depth noise scales with distance, with standard deviations of approximately $0.7$--$0.8\%$ of the measured depth~\cite{Khoshelham2012}, corresponding to $3$--$6\,\text{mm}$ at the manipulation distances considered, motivating our choice of $\tau$.

The friction cone at the contact points is discretized into $N_{dir} = 8$ directions. In the optimized UNCLE-Grasp variant, this is reduced to $N_{dir-opt} = 6$ to improve computational efficiency without degrading performance.

\paragraph{Occlusion-Dependent Confidence Scaling}
The parameter $\alpha \in \{0.0,0.1,0.2,0.3,0.4\}$ controls the severity of
synthetic leaf occlusion and corresponds to the levels described in
Subsection~\ref{Occlusion Levels and Trial Protocol}. To increase risk
aversion as less geometry is observed, the LCB coefficient is varied linearly
with $\alpha$,
\[
z_{\alpha} \in \{0.75,\ 0.88,\ 1.02,\ 1.15,\ 1.28\},
\]
allowing stricter confidence requirements under increasing occlusion. The upper endpoint $z_{\max}=1.28$ was motivated by the 90th percentile of a standard normal distribution and was adopted as a conservative reference value for the most severe occlusion level. However, because the distribution of force-closure scores is not assumed to be Gaussian or probabilistically calibrated, this value does not imply a 90\% confidence guarantee. The lower endpoint $z_{\min}=0.75$, corresponding to approximately the 77th percentile of a standard normal distribution, was used as a permissive reference under full visibility (0\% occlusion) to avoid excessive abstention. Similarly to $z_{\max}$, this percentile also does not imply a calibrated confidence guarantee. The intermediate values were obtained by linear interpolation. Thus, $z_{\alpha}$ is treated as a task-specific risk-sensitivity coefficient, not as a calibrated confidence level.

\paragraph{Uncertainty Threshold Selection}
Due to differences between simulation and the physical robot in sensor noise, depth quantization, and unmodeled physical effects, uncertainty distributions differ, making a single universal threshold impractical. We therefore select thresholds empirically for each domain through preliminary validation and then fix them across all occlusion levels and test trials; therefore, these thresholds are operational cutoffs rather than calibrated probabilities. After preliminary evaluation, we set the global and local uncertainty thresholds to $\delta_{\text{global}} = \delta_{\text{local}} = 0.01$ in simulation. For the physical robot experiments, we apply a stricter threshold of $\delta_{\text{global}} = \delta_{\text{local}} = 0.0037$ to account for increased sensing noise.


\begin{figure}
\centering
\begin{tikzpicture}
\begin{axis}[
    width=0.95\columnwidth,
    height=0.6\columnwidth,
    xlabel={Occlusion level (\%)},
    ylabel={Conditional grasp success rate},
    xmin=0, xmax=90,
    ymin=0, ymax=1.10,
    xtick={0,6.94,28.83,63.12,87.45},
    xticklabels={0, 6.94, 28.83, 63.12, 87.45},
    grid=both,
    legend style={
        at={(0.5,-0.27)},
        anchor=north,
        legend columns=2,
        draw=black,
        font=\scriptsize,
        cells={anchor=west},
        column sep=5pt
    }
]

\addplot[
    color=partialgray,
    mark=o,
    thick,
    error bars/.cd,
        y dir=both,
        y explicit
]
coordinates {
    (0,0.342) +- (0,0.087)
    (6.94,0.348) +- (0,0.110)
    (28.83,0.417) +- (0,0.233)
    (63.12,0.380) +- (0,0.092)
    (87.45,0.337) +- (0,0.093)
};
\addlegendentry{CGNet (Partial)}

\addplot[
    color=geomblue,
    mark=square,
    thick,
    error bars/.cd,
        y dir=both,
        y explicit
]
coordinates {
    (0,0.683) +- (0,0.033)
    (6.94,0.652) +- (0,0.061)
    (28.83,0.708) +- (0,0.100)
    (63.12,0.708) +- (0,0.100)
    (87.45,0.683) +- (0,0.033)
};
\addlegendentry{CGNet+Geom (Partial)}

\addplot[
    color=black,
    mark=triangle,
    thick,
    error bars/.cd,
        y dir=both,
        y explicit
]
coordinates {
    (0,0.495) +- (0,0.106)
    (6.94,0.443) +- (0,0.134)
    (28.83,0.625) +- (0,0.090)
    (63.12,0.655) +- (0,0.157)
    (87.45,0.600) +- (0,0.167)
};
\addlegendentry{Centroid (Completed)}

\addplot[
    color=orange!85!black,
    mark=diamond,
    thick,
    error bars/.cd,
        y dir=both,
        y explicit
]
coordinates {
    (0,0.580) +- (0,0.227)
    (6.94,0.585) +- (0,0.263)
    (28.83,0.647) +- (0,0.197)
    (63.12,0.690) +- (0,0.247)
    (87.45,0.760) +- (0,0.107)
};
\addlegendentry{CGNet (Completed)}

\addplot[
    color=teal!80!black,
    mark=triangle*,
    thick,
    error bars/.cd,
        y dir=both,
        y explicit
]
coordinates {
    (0,0.675) +- (0,0.216)
    (6.94,0.777) +- (0,0.174)
    (28.83,0.710) +- (0,0.196)
    (63.12,0.730) +- (0,0.228)
    (87.45,0.780) +- (0,0.129)
};
\addlegendentry{CGNet+Geom (Completed)}

\addplot[
    color=red!80!black,
    mark=*,
    mark size=2.3pt,
    very thick,
    error bars/.cd,
        y dir=both,
        y explicit,
        error bar style={line width=1.2pt}
]
coordinates {
    (0,0.742) +- (0,0.280)
    (6.94,1.000) +- (0,0.000)
    (28.83,0.750) +- (0,0.250)
    (63.12,0.850) +- (0,0.213)
    (87.45,0.870) +- (0,0.140)
};
\addlegendentry{\textbf{\textit{UNCLE-Grasp (Ours)}}}

\end{axis}
\end{tikzpicture}

\caption{Simulation comparison of five diagnostic ablations and UNCLE-Grasp (Ours) under increasing occlusion. Completion-based methods generally improve grasp robustness over their partial-observation counterparts, while geometric filtering provides additional gains. UNCLE-Grasp achieves the highest conditional grasp success across most occlusion levels through uncertainty-aware filtering and abstention. Values are reported as mean $\pm$ standard deviation over ten trials.}
\label{fig:ablation_plot}
\end{figure}

\section{Results and Discussion}

As discussed in Section~\ref{sec:related_work}, direct quantitative comparison with prior uncertainty-aware grasping methods is not reported, as our approach reasons over learned shape ambiguity under severe occlusion and supports object-level abstention, resulting in a fundamentally different decision scope. Instead, performance is evaluated through controlled ablations designed to isolate the effects of individual pipeline components.

\subsection{Real and Fake Depth Values}


The recorded depth value of the fake strawberry's topmost point is approximately 0.280 m, while the topmost points of the 10 real strawberries were $0.276 \pm 0.008 \text{ m}$. The low standard deviation confirms stable and consistent depth estimation across the real strawberries. No systematic bias was observed between measurements taken on real versus plastic strawberries. This demonstrates that the RGB-D perception pipeline (detection, segmentation, and partial point cloud extraction) operates equivalently for both.

Since the perception pipeline relies on geometric information rather than deformable object modeling, the similarity in depth measurements implies that experimental conclusions derived from plastic strawberries generalize to real strawberries. The only difference arises in the physical manifestation of grasp failure: slippage in plastic strawberries versus deformation in real ones. We accounted for this by recording any instance in which the strawberry is dropped, displaced, or damaged as a grasp failure. Thus, the perception and grasp planning stages of the pipeline remain unaffected, regardless of whether experiments are conducted on real or plastic strawberries.

\subsection{Effect of Occlusion on Uncertainty}

As occlusion increases, the centroid of the partial point cloud is observed to shift by up to $241.7\%$ (Fig.~\ref{fig:occ_levels_centroid_shift}). Since centroid estimates are commonly used to center grasp placement, such shifts can result in misaligned grasps and damage the delicate strawberries. This highlights the unreliability of partial observations for grasp planning and motivates the use of a completion model.

Figure~\ref{fig:completion_uncertainty_bar} shows that MC-dropout completion variability generally increases as less strawberry geometry is observed, indicating greater disagreement among stochastic completion predictions under heavier occlusion. However, this variability measures disagreement in reconstructed geometry rather than grasp feasibility. Different completions may vary geometrically while still supporting similarly feasible grasps. We therefore use completion variability as an uncertainty signal for downstream filtering rather than as a direct criterion for rejecting a strawberry.

\subsection{Selective Execution and Overall Harvesting Performance}
\label{sec:selective_execution}

Tables~\ref{tab:success_over_total}, \ref{tab:attempt_results}, and \ref{tab:occlusion_results} show that UNCLE-Grasp improves success over attempted strawberries primarily through selective abstention. At several occlusion levels, its attempt rate is lower than that of the completed baselines, so the higher conditional success rate does not always translate into a higher overall success rate. For example, at 6.94\% simulated occlusion, it attempts only 0.280 of the available strawberries, but every executed attempt succeeds, yielding a conditional success rate of 1.000 and an overall success rate of 0.280. This illustrates the intended effect of uncertainty filtering: ambiguous targets are rejected to improve the reliability of executed grasps, at the cost of attempting fewer strawberries. A similar trade-off appears at 28.83\% occlusion, when the overall success rate was 0.200 and we attempted 0.300 of the strawberries, we got a conditional success rate of 0.750, meaning that we rejected most of the risky strawberries that lowered the overall success rate.

At the highest simulated occlusion, however, UNCLE-Grasp attempts 0.860 of the strawberries, close to the 0.880 attempt rate of CGNet+Geom (Completed), while improving success over attempts from 0.780 to 0.870 and overall success from 0.680 to 0.740. This higher attempt rate does not imply lower uncertainty. Although completion variability increases with occlusion, the uncertainty filters use fixed thresholds, so an individual target may remain below the rejection threshold. Moreover, the final decision depends on the LCB of grasp feasibility across completions rather than geometric variability alone. As shown in Fig.~\ref{fig:qualitative_occlusions}, severe occlusion leaves only sparse observed geometry, causing the completion model to rely more strongly on its learned shape prior. This can produce canonical completed shapes that remain favorable for grasp generation and filtering. Consequently, geometrically different completions may still provide sufficiently consistent positive grasp support for the LCB to remain above zero, resulting in an attempted grasp. Attempt rate is therefore not expected to decrease monotonically with occlusion.

On the physical robot, the same trade-off is more pronounced, with high conditional success often accompanied by lower attempt rates. Across most reported settings, UNCLE-Grasp exhibits lower attempt rates than the completed baselines because its uncertainty-aware filtering and abstention mechanisms deliberately reject strawberries considered unreliable for execution. Importantly, the increase in conditional grasp success suggests that UNCLE-Grasp is primarily rejecting unreliable targets rather than safe ones. If safe strawberries were rejected instead, the success rate of the remaining grasp attempts would not increase as much. Overall, the results show that the main benefit of UNCLE-Grasp is risk-aware grasp decision-making: it improves the reliability of executed grasps, although this can reduce harvesting yield when many uncertain targets are rejected.

\begin{figure}
    \centering
    \includegraphics[width=\linewidth]{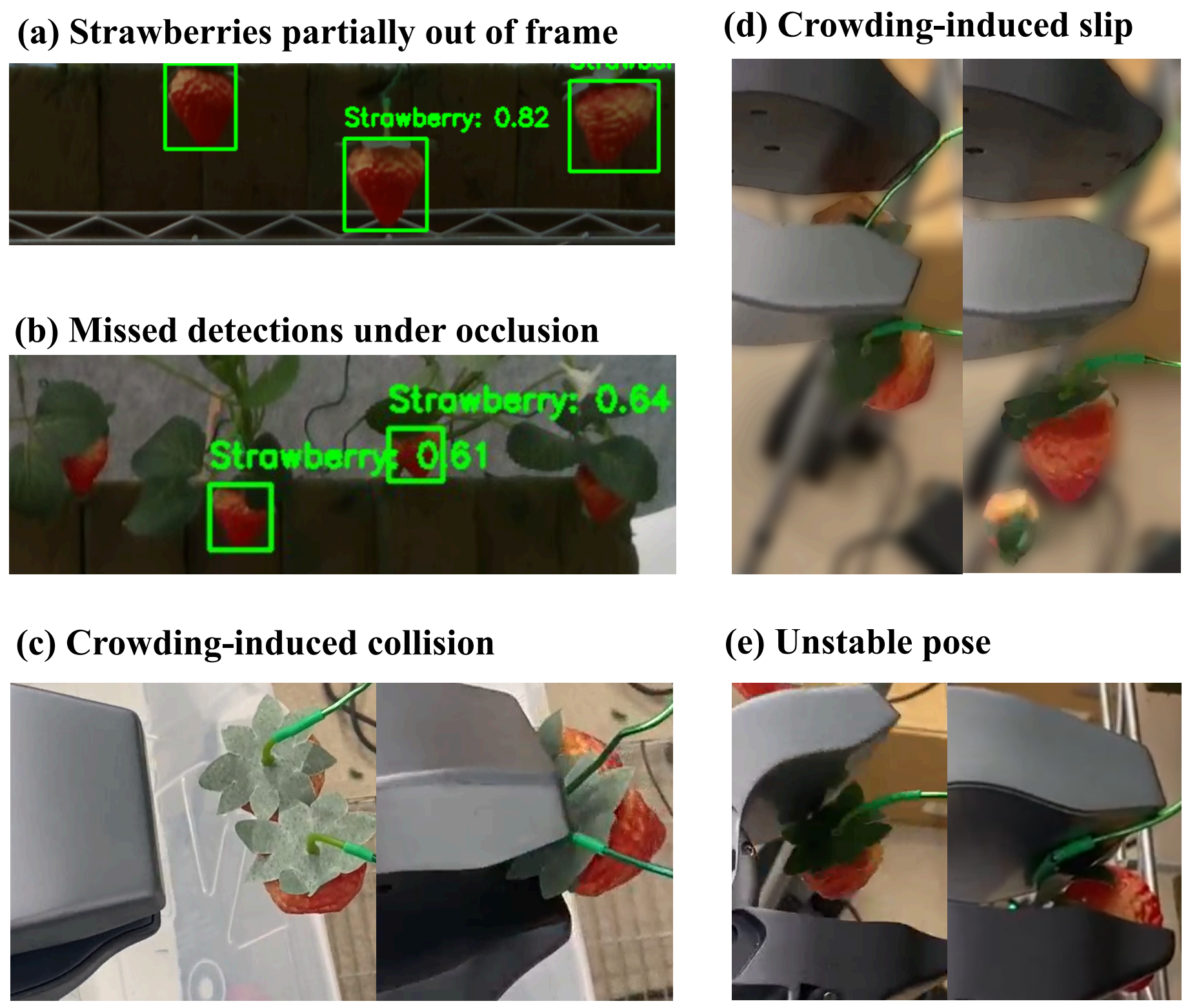}
    \caption{Failure modes observed when executing UNCLE-Grasp.
\textbf{Perception: (a-b) }Partial observability due to limited field of view and occlusion leads to undetected strawberries.
\textbf{Physical interaction: (c-d) }During end-effector approach, a crowded plant with unmodeled neighboring strawberries induces slippage and collisions.
\textbf{Geometric uncertainty: (e) }Pose uncertainty results in unstable grasp execution despite apparent grasp feasibility.
}
    \label{fig:failure}
\end{figure}

\subsection{Pipeline Ablations}

Since UNCLE-Grasp abstains from executing grasps on high-risk strawberries, the conditional success rate best reflects the reliability of the grasps that are actually attempted. We therefore use conditional grasp success as the primary metric for evaluating grasp quality. Figure~\ref{fig:ablation_plot} reports all six ablations, allowing us to isolate the contribution of shape completion, learned grasp generation, geometric filtering, and uncertainty-aware grasp selection.

\subsubsection{Effect of Shape Completion}

Applying CGNet directly to partial point clouds results in poor performance, particularly as occlusion becomes severe. At the highest simulated occlusion level, CGNet (Partial) achieves a grasp success rate of only 0.337, whereas CGNet (Completed) reaches 0.760. The same trend remains when geometric filtering is applied: at approximately 63.12\% occlusion, CGNet+Geom (Partial) achieves 0.633, while CGNet+Geom (Completed) improves to 0.730. These results show that recovering the occluded geometry provides substantially more useful information for grasp generation than operating directly on incomplete observations.

The apparent improvement of some completion-based variants at higher occlusion should not be interpreted as evidence that occlusion itself benefits grasping. As less geometry becomes visible, the completion model relies more strongly on its learned shape prior and can produce smoother, more canonical strawberry shapes that are easier for the grasp generator and geometric filters to process, even if those completions are not necessarily more accurate.

\subsubsection{Effect of Geometric Filtering}

Geometric filtering improves grasp success by rejecting physically infeasible grasp candidates. For partial observations, the effect is substantial: at 0\% occlusion, CGNet (Partial) achieves a success rate of 0.342, while CGNet+Geom (Partial) reaches 0.683. At approximately 63.12\% occlusion, the corresponding improvement is from 0.380 to 0.633. This indicates that physically grounded constraints can compensate for some of the weaknesses of grasp generation from incomplete geometry.

The same benefit is observed for completed point clouds. At the highest simulated occlusion level, CGNet (Completed) achieves a success rate of 0.760, while CGNet+Geom (Completed) improves to 0.780. Although the gain is smaller once completion is available, the consistent improvement shows that geometric filtering remains useful for removing infeasible grasps. Its effectiveness nevertheless saturates under severe ambiguity because the filtering remains deterministic and depends on the accuracy of the estimated geometry.

\subsubsection{Effect of Learned Grasp Generation}

The completed-shape ablations also demonstrate the benefit of generating grasps using CGNet rather than relying on simple centroid-based placement. Centroid (Completed) provides a reasonable baseline after shape completion, but its performance degrades relative to learned grasp generation as occlusion becomes more challenging. At the highest simulated occlusion level, Centroid (Completed) achieves a success rate of 0.600, compared with 0.760 for CGNet (Completed) and 0.780 for CGNet+Geom (Completed). Similarly, at 28.83\% occlusion, the combination of learned grasp generation and geometric filtering achieves 0.710 compared with 0.625 for Centroid (Completed). These results indicate that shape completion alone is insufficient: the recovered geometry must also be paired with an effective grasp-selection mechanism.

Together, the diagnostic ablations establish CGNet+Geom (Completed) as the strongest deterministic baseline. Shape completion provides the largest gains under severe occlusion, learned grasp generation improves upon simple centroid-based placement, and geometric filtering further removes physically infeasible candidates. However, none of these components explicitly accounts for uncertainty in the completed geometry.

\begin{table}
\caption{Effect of fruit-aware planning in a simplified 0\% occlusion
setting with centroid-based grasping.}
\label{tab:success_and_hit_rate_strawberry}
\centering
\begin{tabular}{ccc}
\toprule
\makecell{\textbf{Fruit-Aware} \\ \textbf{Planning}}
& \makecell{\textbf{Grasp Success} \\ \textbf{Rate} $\uparrow$}
& \makecell{\textbf{Obstacle Hit} \\ \textbf{Rate} $\downarrow$} \\
\midrule
\xmark & 0.723 & 0.433 \\
\cmark & \textbf{0.896} & \textbf{0.140} \\
\bottomrule
\end{tabular}
\end{table}

\subsubsection{Effect of Uncertainty-Aware Grasp Selection}

Despite increasing completion uncertainty under severe occlusion (Fig.~\ref{fig:completion_uncertainty_bar}), UNCLE-Grasp consistently achieves the highest conditional grasp success across occlusion levels. At the highest simulated occlusion level, success increases from 0.780 for CGNet+Geom (Completed) to 0.870 with UNCLE-Grasp. On the physical robot, UNCLE-Grasp achieves a success rate of 0.800 under the corresponding synthetic occlusion setting, compared with 0.483 for the reference baseline. Under approximately 70\% real leaf occlusion, it improves grasp success from 0.367 for CGNet+Geom (Completed) to 0.517.

These results show that completion and deterministic geometric filtering alone are insufficient when the visible geometry becomes highly ambiguous. Geometric filtering removes grasps that are physically infeasible with respect to the estimated shape, but it cannot determine whether that shape estimate is itself reliable. UNCLE-Grasp addresses this limitation through uncertainty-aware LCB selection and abstention, suppressing grasp attempts when the completed geometry is insufficiently reliable. Robust grasping under heavy occlusion therefore arises from the complementary use of shape completion, learned grasp generation, geometric filtering, and uncertainty-aware selection.

\subsection{Failure Case Mitigation}
\label{app:failure_cases}

We analyze representative failure cases observed during physical execution to characterize the operating scope of UNCLE-Grasp and distinguish grasp-feasibility prediction from challenges arising during downstream robotic execution. As summarized in Fig.~\ref{fig:failure}, the observed failures are primarily associated with partial observability, geometric uncertainty, and physical interactions that emerge as the gripper approaches the strawberries. These cases do not necessarily indicate incorrect object-level grasp-feasibility decisions; rather, they highlight execution-stage factors that are not fully observable at the time of grasp assessment or captured by the current perception and manipulation pipeline. This distinction clarifies the role of UNCLE-Grasp as an object-level grasp-feasibility reasoning method while identifying complementary execution-aware modeling as a promising direction for further improving end-to-end physical grasping performance.

\subsubsection{Partial Observability}
As shown in Fig.~\ref{fig:failure}(a-b), limited camera field of view and leaf occlusion often result in strawberries that are partially out of frame or entirely undetected. While a target strawberry may be correctly identified, neighboring fruit remain unmodeled, leading to incomplete scene understanding. These cases illustrate how perception-level partial observability propagates downstream and compromises safe execution in cluttered plants.

\subsubsection{Crowding-Induced Interaction During Approach}
Fig.~\ref{fig:failure}(c-d) show failures that occur during the end-effector approach in densely cluttered strawberry plants. Unmodeled neighboring fruit induce unintended contacts, resulting in collisions or slippage even when the selected grasp on the target strawberry is geometrically feasible. These failures highlight the insufficiency of grasp-level reasoning alone and motivate path planning strategies that explicitly account for nearby fruit during approach.

\subsubsection{Unstable Execution}
As illustrated in Fig.~\ref{fig:failure}(e), pose uncertainty in the estimated strawberry geometry can lead to unstable grasp execution despite apparent grasp feasibility. Small orientation or position errors result in poor contact configurations that cause grasp instability and fruit damage. This failure mode emphasizes the need to reason about geometric uncertainty beyond binary grasp feasibility.

\subsection{Mitigating Neighboring Fruit Failures}
The 5 failure cases discussed show that reliable object-level grasp selection alone is not sufficient for complete autonomous harvesting in densely cluttered plants. UNCLE-Grasp operates after a target strawberry has been detected and segmented, and its contribution is limited to grasp filtering and the attempt-or-abstain decision for that target. It does not address missed detections, complete scene reconstruction, or collision-aware approach planning.

To illustrate the effect of incorporating neighboring fruit during motion planning, we evaluate a fruit-aware planning strategy in a simplified setting with centroid-based grasping and no synthetic occlusion. Non-target strawberries are modeled as obstacles during execution. As shown in Table~\ref{tab:success_and_hit_rate_strawberry}, the baseline operates on partial point clouds without fruit-aware planning, whereas the second setting uses completed point clouds with fruit-aware planning enabled. The obstacle hit rate decreases from 0.433 to 0.140 (a 67.8\% reduction), while grasp success increases from 0.723 to 0.896. Because the two settings differ in both scene representation (partial and completed point clouds) and motion planning (fruit-aware planning), the observed improvements should be interpreted as evidence for the benefit of richer scene-level reasoning as a whole, rather than as an isolated measure of the contribution of fruit-aware planning or UNCLE-Grasp itself.

\section{Limitations and Future Work}

\label{inference}
\begin{table} 
\centering
\caption{Average inference time per module.}
\label{tab:timing}
\scriptsize
\resizebox{\linewidth}{!}{%

\begin{tabular}{l l c}
\hline
\textbf{Method} & \textbf{Module} & \textbf{Time (s)} \\
\hline 
\multirow{4}{*}{Shared} 
& YOLO detection (per frame) & 0.30 \\
& SAM segmentation (per frame) & 0.22 \\
& Completion (per strawberry) & 0.79 \\
& CGNet (per strawberry) & 1.57 \\
\hline \hline

\multirow{2}{*}{CGNet (Completed)}
& Grasp selection and filtering & 0.01 \\
& Total (per strawberry) & 2.48 \\
\hline \hline

\multirow{2}{*}{CGNet+Geom (Completed)}
& Filtering and selection & 9.71 \\
& Total (per strawberry) & 11.86 \\
\hline \hline

\multirow{6}{*}{UNCLE-Grasp \textit{(Ours)}}
& MC completion (20 forward passes) & 15.86 \\
& CGNet (20 forward passes) & 37.72 \\ \cline{2-3}
& LCB aggregation (original) & 208.17 \\
& Total (original, per strawberry) & 230 \\ \cline{2-3} 
& \textbf{LCB aggregation (optimized)} & \textbf{1.21} \\
& \textbf{Total (optimized, per strawberry)} & \textbf{57.53} \\
\hline
\end{tabular}%
}
\end{table}

The proposed pipeline is a research prototype designed to evaluate uncertainty-aware grasp selection rather than a production-ready harvesting system. Although real-time operation is not the primary objective of the present study, computational efficiency remains an important limitation. As shown in Table~\ref{tab:timing}, the optimized UNCLE-Grasp pipeline requires approximately 57.53~s per strawberry, compared with 2.48~s for CGNet (Completed) and 11.86~s for CGNet+Geom (Completed). Most of this cost arises from repeated Monte Carlo shape completions and grasp evaluations. Optimizing the LCB aggregation reduces its runtime from 208.17~s to 1.21~s and provides an approximately $4\times$ reduction in total inference time; however, the pipeline remains substantially slower than the deterministic baselines. Furthermore, its improved success over attempted grasps is partly achieved by abstaining from uncertain targets, which may reduce the number of strawberries harvested per unit time. The present results should therefore be interpreted as demonstrating the reliability-throughput trade-off introduced by uncertainty-aware selective execution, rather than establishing readiness for real-time commercial deployment.

The current uncertainty estimator captures only dropout-induced variability in the completion model, which we interpret as an approximation of epistemic uncertainty. Its ability to detect out-of-distribution inputs is not evaluated, and it should therefore not be interpreted as a validated OOD-detection mechanism. It does not explicitly parameterize aleatoric uncertainty from RGB-D measurement noise or irreducible ambiguity in the mapping from a partial observation to a complete shape. These uncertainty sources are therefore not disentangled. In particular, low variability across dropout samples does not guarantee that the hidden geometry is unambiguous, because different model samples may agree on the same completion despite unresolved conditional ambiguity. Future work will investigate explicit observation-noise models and conditional generative completion methods that can represent and propagate epistemic and aleatoric uncertainty separately.

In addition, the system relies on single-view RGB-D observations; under extreme occlusion, insufficient visible geometry can still lead to ambiguous completions. Moreoever, grasp execution is evaluated using a fixed gripper model, which does not fully capture soft fruit compliance, friction variations, or dynamic effects. UNCLE-Grasp does not guarantee the success of any particular grasp since it is inherently probabilistic. It instead indicates that with high confidence, at least one feasible grasp exists for the object despite geometric uncertainty. 

Future work can further reduce execution time by optimizing the remaining computational bottlenecks and parallelizing independent stages of the pipeline, particularly the Monte Carlo shape-completion and grasp-evaluation operations. Furthermore, it would be useful to work on integrating active perception and next-best-view strategies to reduce uncertainty prior to grasping and to extend the approach to multi-view and multi-object harvesting. Additional strategies to further improve robustness include prioritizing the most visible or least uncertain fruits, whose removal could help clear occlusions, and actively manipulating leaves to reduce clutter. In principle, a failed attempt could be followed by selecting the next best-scoring grasp; however, we leave such multi-attempt strategies to future work.

The detection module is modular and extensible. The same approach can be generalized to other fruits by retraining YOLO on the corresponding dataset. Once a fruit is detected, the pipeline retrieves the partial point cloud within the bounding box from the aligned depth image. Since this step depends solely on geometric segmentation within the detected region, the resulting partial point cloud structure would remain identical in form for other fruit types, provided an accurate detector is available.

\section{Conclusion}
This study shows that robust manipulation under partial observability requires
reasoning not only about the quality of a predicted action, but also about the
reliability of the incomplete geometric representation on which that action is
based. UNCLE-Grasp addresses this problem by propagating variability across
stochastic shape completions through physically grounded grasp evaluation to
an object-level attempt-or-abstain decision.

The results highlight a broader principle for risk-aware automation: when
perception is ambiguous, always acting can be less desirable than selectively
abstaining. This improves the reliability of executed actions, but introduces
a trade-off with attempt rate, harvesting yield, and computational throughput. Accordingly, uncertainty-aware systems should be evaluated using both conditional success and overall grasp success across all target opportunities, including those on which the system abstains.

More generally, the study suggests that uncertainty is most useful when it is
connected to an explicit downstream decision rather than reported only as a
perception score. At the same time, the observed failures show that reliable
object-level decisions cannot compensate for missed detections, incomplete
scene representations, or collision-prone approach motions. Risk-aware grasp
selection should therefore be viewed as one component of a broader harvesting
system that combines uncertainty estimation, active perception, scene-level
reasoning, and collision-aware planning.


\appendix
\section{Algorithm}
\label{app:algorithm}

The proposed UNCLE-Grasp pipeline is summarized in
Algorithms~\ref{alg:strawberry_pipeline_1}
and~\ref{alg:strawberry_pipeline_2}.

\begin{algorithm} 
\caption{UNCLE-Grasp pipeline: preprocessing, completion, and deterministic baselines}
\label{alg:strawberry_pipeline_1}
\small

\begin{algorithmic}[1]

\Statex \textbf{Input:}
\Statex \hspace{\algorithmicindent}
RGB-D frame $(I_{\mathrm{RGB}}, I_D)$ and camera model $\Pi$
\Statex \hspace{\algorithmicindent}
Completion network \textsc{PointAttN} 
\Statex \hspace{\algorithmicindent} Grasp network \textsc{CGNet}
\Statex \hspace{\algorithmicindent}
Mode $\in \{$\textsc{CGNet(Completed)}, \textsc{CGNet+Geometric(Completed)}, \textsc{UNCLE-Grasp}$\}$
\Statex \hspace{\algorithmicindent}
MC samples $K$ and thresholds
$\theta_{\mathrm{dot}},\theta_{\mathrm{vert}},\tau$
\Statex \hspace{\algorithmicindent}
Uncertainty thresholds
$\delta_{\mathrm{global}},\delta_{\mathrm{local}}$
and confidence factor $z_{\alpha}$

\Statex \textbf{Output:}
\Statex \hspace{\algorithmicindent}
\textsc{Attempt} or \textsc{Abstain}

\Statex
\Statex \textbf{Preprocessing and Segmentation}

\State $\mathcal{P}_{\mathrm{scene}}
\gets
\textsc{DepthFilterProject}
(I_{\mathrm{RGB}},I_D,\Pi)$

\State $\mathcal{P}^{\mathrm{all}}_{\mathrm{denoised}}
\gets
\textsc{SegmentAndDenoise}
(\mathcal{P}_{\mathrm{scene}})$

\ForAll{$\mathcal{P}_{\mathrm{denoised}}
\in
\mathcal{P}^{\mathrm{all}}_{\mathrm{denoised}}$}

    \Statex \textbf{Shape Completion}

    \If{$\mathrm{mode}\neq\textsc{UNCLE-Grasp}$}

        \State $\mathcal{P}_{\mathrm{completed}}
        \gets
        \textsc{PointAttN}
        (\mathcal{P}_{\mathrm{denoised}})$

    \Else

        \State $\{
        \mathcal{P}_{\mathrm{completed}}^{(k)}
        \}_{k=1}^{K}
        \gets
        \textsc{PointAttN}
        (\mathcal{P}_{\mathrm{denoised}})$

        \Statex using MC dropout

    \EndIf

    \Statex \textbf{Grasp Generation}

    \If{$\mathrm{mode}\neq\textsc{UNCLE-Grasp}$}

        \State $\{
        \mathbf{G}_i,s_i
        \}_{i=1}^{M}
        \gets
        \textsc{CGNet}
        (\mathcal{P}_{\mathrm{completed}})$

    \Else

        \For{$k=1,\dots,K$}

            \State $\{
            \mathbf{G}_i^{(k)},s_i^{(k)}
            \}_{i=1}^{M}
            \gets
            \textsc{CGNet}
            (\mathcal{P}_{\mathrm{completed}}^{(k)})$

        \EndFor

    \EndIf

    \Statex \textbf{Deterministic Baselines}

    \If{$\mathrm{mode}=\textsc{CGNet(Completed)}$}

        \State Execute the grasp with the highest $s_i$
        \State \Return \textsc{Attempt}

    \EndIf

    \If{$\mathrm{mode}
    =
    \textsc{CGNet+Geometric(Completed)}$}

        \State $\{
        \mathbf{G}_j,s_j
        \}_{j=1}^{M'}
        \gets$ $\textsc{GeomFilter}
        \left(
        \{\mathbf{G}_i,s_i\}_{i=1}^{M},
        \mathcal{P}_{\mathrm{completed}}
        \right)$

        \If{$M'=0$}

            \State \Return \textsc{Abstain}

        \Else

            \State Execute grasp
            $\displaystyle\arg\max_j s_j$

            \State \Return \textsc{Attempt}

        \EndIf

    \EndIf

\algstore{unclegrasp}

\end{algorithmic}
\end{algorithm}

\begin{algorithm} 
\caption{UNCLE-Grasp pipeline: uncertainty-aware grasp decision}
\label{alg:strawberry_pipeline_2}
\small

\begin{algorithmic}[1]

\algrestore{unclegrasp}

    \Statex \textbf{UNCLE-Grasp Uncertainty-Aware Mode}

    \If{$\mathrm{mode}=\textsc{UNCLE-Grasp}$}

        \If{$
        \textsc{GlobalUnc}
        \left(
        \{
        \mathcal{P}_{\mathrm{completed}}^{(k)}
        \}_{k=1}^{K}
        \right)
        >
        \delta_{\mathrm{global}}
        $}

            \State \Return \textsc{Abstain}

        \EndIf

        \For{$k=1,\dots,K$}

            \State $\{
            \mathbf{G}_j^{(k)},s_j^{(k)}
            \}_{j=1}^{M'_k}
            \gets$ $\textsc{LocalUnc+GeomFilter}
            \left(
            \{
            \mathbf{G}_i^{(k)},s_i^{(k)}
            \}_{i=1}^{M},
            \mathcal{P}_{\mathrm{completed}}^{(k)}
            \right)$

            \If{$M'_k=0$}

                \State $\epsilon_k\gets0$

            \Else

                \State Estimate contacts for
                $\{
                \mathbf{G}_j^{(k)}
                \}_{j=1}^{M'_k}$


                \State Construct the combined grasp wrench space
                from $\{
                \mathbf{G}_j^{(k)}
                \}_{j=1}^{M'_k}$
            
                \State $\epsilon_k
                \gets
                \epsilon
                \left(
                \mathcal{G}_k
                \right)$

            \EndIf

        \EndFor

        \State Compute $\bar{\epsilon}$ and $\sigma_{\epsilon}$

        \State $\mathrm{LCB}
        \gets
        \bar{\epsilon}
        -
        z_{\alpha}\sigma_{\epsilon}$

        \If{$\mathrm{LCB}\leq0$}

            \State \Return \textsc{Abstain}

        \Else

            \State \Return \textsc{Attempt}
            \Comment{A feasible grasp exists}

        \EndIf

    \EndIf

    \State \Return \textsc{Abstain}

\EndFor

\end{algorithmic}
\end{algorithm}

\section{Data Collection}
We used a mixture of simulation and physical robot data to train our completion model. For simulation data, we utilized NVIDIA Isaac Sim to create a virtual strawberry field with realistic occlusions and simulated an Intel RealSense D435i RGB-D camera from the robot's perspective to capture partial point clouds of strawberries. This setup also provided complete point clouds from the simulator, which served as the ground truth for training and evaluation. Since no large-scale real-world dataset of occluded/complete strawberry pairs exists, the complete shapes were taken directly from Isaac Sim as ground truth in simulation. 

For physical robot data collection, we captured RGB-D data using an Intel RealSense D435i camera in our lab environment, designed to replicate an indoor greenhouse strawberry plantation. The lab setup included five hanging strawberries with varied arrangements to simulate realistic harvesting conditions. The captured RGB-D data was processed to generate partial point clouds, which were manually annotated by aligning them with an approximated 3D CAD model. Both the simulation and physical robot environments, including the camera perspectives and strawberry field layouts, are shown in Fig. \ref{fig:sim_real_setup_labels}.

\section{RGB-D Instance Point-Cloud Preprocessing}
\label{app:preprocessing}

This appendix provides implementation details for the standard RGB-D
preprocessing operations summarized in Section~\ref{sec:preprocessing}.
The preprocessing is applied independently to each RGB-D frame before shape
completion and grasp generation.

\subsection{Depth Filtering}

Let $I_{\mathrm{RGB}}$ and $I_D$ denote the aligned RGB and depth images,
respectively. To suppress isolated depth noise while preserving depth
discontinuities, we apply a $5\times5$ median filter to the depth image:
\begin{equation}
I_{D,\mathrm{filtered}}(u,v)
=
\operatorname{median}
\left\{
I_D(u',v')
\;\middle|\;
(u',v')\in\mathcal{N}_5(u,v)
\right\},
\label{eq:appendix_median_filter}
\end{equation}
where $(u,v)$ denotes a pixel location and
$\mathcal{N}_5(u,v)$ is the $5\times5$ neighborhood centered at that
location. Pixels containing invalid depth measurements, including zero,
NaN, or infinite values, are discarded before constructing the point cloud.

\subsection{RGB-D Back-Projection}

Let the calibrated camera-intrinsic matrix be
\begin{equation}
K =
\begin{bmatrix}
f_x & 0   & c_x \\
0   & f_y & c_y \\
0   & 0   & 1
\end{bmatrix},
\label{eq:appendix_intrinsics}
\end{equation}
where $f_x$ and $f_y$ are the focal lengths and $(c_x,c_y)$ is the
principal point. For a valid depth value
$d(u,v)=I_{D,\mathrm{filtered}}(u,v)$, the corresponding 3D point in the
camera coordinate frame is obtained by
\begin{equation}
p(u,v)
=
d(u,v)
K^{-1}
\begin{bmatrix}
u\\
v\\
1
\end{bmatrix}.
\label{eq:appendix_backprojection}
\end{equation}
Equivalently,
\begin{equation}
p(u,v)
=
\begin{bmatrix}
\displaystyle \frac{(u-c_x)d(u,v)}{f_x}\\[4pt]
\displaystyle \frac{(v-c_y)d(u,v)}{f_y}\\[4pt]
d(u,v)
\end{bmatrix}.
\label{eq:appendix_backprojection_expanded}
\end{equation}
The scene-level point cloud is therefore
\begin{equation}
\mathcal{P}_{\mathrm{scene}}
=
\left\{
p(u,v)
\;\middle|\;
d(u,v)\text{ is valid}
\right\}
\subset\mathbb{R}^{3}.
\label{eq:appendix_scene_cloud}
\end{equation}

The RGB image is used for detection and segmentation, while the aligned
depth image and camera intrinsics are used for geometric back-projection.
Because the RGB and depth images are spatially aligned, an image-space
instance mask can be associated directly with the corresponding depth
pixels.

\subsection{Voxel-Grid Downsampling}

To reduce point-cloud density and computational cost, we apply voxel-grid
downsampling with voxel size $v_s$. Let $V_i$ denote the set of points
assigned to voxel index $i$:
\begin{equation}
V_i
=
\left\{
p_j\in\mathcal{P}_{\mathrm{scene}}
\;\middle|\;
\left\lfloor
\frac{p_j}{v_s}
\right\rfloor=i
\right\},
\label{eq:appendix_voxel_assignment}
\end{equation}
where division and the floor operation are applied element-wise.

For each retained voxel, its representative point is the centroid
\begin{equation}
c_i
=
\frac{1}{|V_i|}
\sum_{p_j\in V_i}p_j,
\qquad
|V_i|\geq n_{\min},
\label{eq:appendix_voxel_centroid}
\end{equation}
where $n_{\min}$ is the minimum point-support threshold. Voxels satisfying
$|V_i|<n_{\min}$ are discarded. The downsampled scene point cloud is
\begin{equation}
\mathcal{P}_{\mathrm{voxel}}
=
\left\{
c_i
\;\middle|\;
|V_i|\geq n_{\min}
\right\}.
\label{eq:appendix_voxel_cloud}
\end{equation}

\subsection{Strawberry Detection and Instance Extraction}

Strawberry instances are first detected in $I_{\mathrm{RGB}}$ using
YOLOv8, and each predicted bounding box is refined into a binary instance
mask using SAM2. Let
\begin{equation}
\mathcal{M}
=
\left\{
m_r
\right\}_{r=1}^{R}
\label{eq:appendix_instance_masks}
\end{equation}
denote the set of masks for the $R$ detected strawberries.

For a 3D point $p=[x,y,z]^\top$, its projection into the image plane is
\begin{equation}
\pi_K(p)
=
\begin{bmatrix}
f_x x/z+c_x\\
f_y y/z+c_y
\end{bmatrix}.
\label{eq:appendix_projection}
\end{equation}
The point cloud associated with the $r$-th strawberry is extracted as
\begin{equation}
\mathcal{P}^{(r)}
=
\left\{
p\in\mathcal{P}_{\mathrm{voxel}}
\;\middle|\;
\pi_K(p)\in m_r
\right\}.
\label{eq:appendix_instance_extraction}
\end{equation}
This operation retains only the 3D points whose image projections lie within
the corresponding strawberry mask.

\subsection{Outlier Removal}

Residual isolated measurements are removed from each instance-level point
cloud using a statistical outlier filter. Denoting this operation by
$\mathcal{O}(\cdot)$, the final partial and denoised point cloud for the
$r$-th strawberry is
\begin{equation}
\mathcal{P}^{(r)}_{\mathrm{denoised}}
=
\mathcal{O}
\left(
\mathcal{P}^{(r)}
\right).
\label{eq:appendix_outlier_removal}
\end{equation}
The resulting point cloud is used as the input to the point-cloud completion
network:
\begin{equation}
\mathcal{P}_{\mathrm{denoised}}
=
\mathcal{P}^{(r)}_{\mathrm{denoised}}.
\label{eq:appendix_preprocessing_output}
\end{equation}
The same preprocessing sequence is applied independently to every detected
strawberry instance.

\paragraph{Point Cloud Preprocessing}
For voxel-based downsampling, we set the voxel size to $v_s = 0.05\,\text{m}$ and the minimum point-support threshold to $n_{\min} = 30$. These values were selected empirically via trial-and-error based on the characteristics of our dataset, balancing noise suppression and geometric detail preservation.

\section{Synthetic Leaf Occlusion Modeling}
\label{leaf_model}
\paragraph{Leaf Placement}
A synthetic leaf is attached to one of the four lateral sides of the strawberry. The attachment axis $s \in \{x,y\}$ and sign $\sigma \in \{-1,+1\}$ are sampled uniformly at random. The leaf center $\mathbf{c}_{\text{leaf}}$ is positioned at the corresponding boundary of the bounding box:
\begin{equation}
\mathbf{c}_{\text{leaf}} =
\begin{cases}
(x_{\max}, y_c, z_c), & \text{if } s=x,\ \sigma=+1 \\
(x_{\min}, y_c, z_c), & \text{if } s=x,\ \sigma=-1 \\
(x_c, y_{\max}, z_c), & \text{if } s=y,\ \sigma=+1 \\
(x_c, y_{\min}, z_c), & \text{if } s=y,\ \sigma=-1
\end{cases},
\end{equation}
where $(x_c, y_c, z_c)$ denotes the bounding-box center.

The leaf normal $\mathbf{n}$ points inward toward the object center, and two orthonormal vectors $\mathbf{a}_1$ and $\mathbf{a}_2$ span the leaf plane.

\paragraph{Leaf Geometry}
The leaf footprint is modeled as an ellipse lying in the plane spanned by $\mathbf{a}_1$ and $\mathbf{a}_2$. We introduce a scalar occlusion parameter $\alpha$ that controls the relative size of the synthetic leaf to indicate the occlusion severity.
The major and minor axes of the ellipse are defined as
\begin{equation}
a = \alpha \cdot d, \qquad b = \alpha_{max} a.
\end{equation}

For each point $\mathbf{p}_i$, we compute its coordinates in the leaf reference frame:
\begin{align}
\mathbf{r}_i &= \mathbf{p}_i - \mathbf{c}_{\text{leaf}}, \\
u_i &= \mathbf{r}_i \cdot \mathbf{a}_1, \quad
v_i = \mathbf{r}_i \cdot \mathbf{a}_2, \quad
d_i = \mathbf{r}_i \cdot \mathbf{n}.
\end{align}

A point lies within the elliptical footprint if
\begin{equation}
\left(\frac{u_i}{a}\right)^2 + \left(\frac{v_i}{b}\right)^2 \leq 1.
\end{equation}

\paragraph{Leaf Thickness}
To model the finite thickness of real leaves, a point is considered occluded only if it also lies within a slab of thickness $t_{\text{leaf}}$ along the normal direction:
\begin{equation}
|d_i| \leq t_{\text{leaf}}.
\end{equation}

\paragraph{Occlusion Mask}
A point $\mathbf{p}_i$ is removed from the point cloud if it satisfies both of the following footprint and thickness constraints:
\begin{equation}
\mathbf{p}_i \text{ occluded } \iff
\left(\frac{u_i}{a}\right)^2 + \left(\frac{v_i}{b}\right)^2 \leq 1
\;\land\;
|d_i| \leq t_{\text{leaf}}.
\label{eq:leaf_mask}
\end{equation}

The resulting occluded point cloud is given by
\begin{equation}
\mathcal{P}_{\text{occ}} =
\{\mathbf{p}_i \in \mathcal{P} \mid \mathbf{p}_i \text{ not occluded}\}.
\end{equation}

\paragraph{Resulting Occlusion Levels}
\label{Resulting Occlusion Levels}
While $\alpha$ controls leaf geometry, the resulting fraction of removed points depends on object shape and placement, producing a range of realistic occlusion levels from fully visible fruit to extreme coverage. Increasing occlusion can substantially shift the partial point cloud centroid (Fig.~\ref{fig:occ_levels_centroid_shift}), motivating the use of shape completion rather than grasping directly from partial observations.

\section{Training Details}
The completion model was trained on 420 combined simulation and physical robot partial point cloud samples to enhance robustness and bridge the sim-to-real gap. All 67 validation and 50 test samples consist exclusively of physical robot data, ensuring evaluation under realistic sensing conditions.

Training used 1748 as a fixed seed, batch size of 16 for 400 epochs, with each point cloud uniformly sampled to 2,048 points. We employed the Adam optimizer with an initial learning rate of $1\times10^{-4}$, no weight decay, and applied a learning rate decay of 0.7 every 40 epochs. 

At inference, the model predicts complete strawberry shapes directly from partial scans. Ground truth is not used for prediction but solely for evaluation (e.g., Chamfer Distance). The validation and test sets consist exclusively of physical robot data, ensuring that performance reflects realistic deployment scenarios. While physical robot annotations inevitably include minor manual imperfections, exposure to such noise during training improves the model’s robustness in practice.

For grasp pose generation, we employ the publicly available CGNet model in an off-the-shelf manner. At inference, CGNet takes the completed point cloud predicted by our shape completion network and outputs a set of candidate grasp orientations and approach directions.

\section{Vision Modules}
We trained a YOLOv8 \cite{yolov8_ultralytics} model for strawberry detection using RGB images collected from both the simulation and physical robot environments. The training dataset was augmented with variations in lighting, occlusion, and strawberry orientations to improve generalization.

Although SAM2 is computationally expensive, applying it only within YOLO-predicted bounding boxes keeps the pipeline practical, which is sufficient for mobile harvesting robots where precision outweighs high frame rate. For each detected strawberry, the 2D mask is projected onto the 3D point cloud to extract the fruit-specific point cloud, $\mathcal{P}$, which we call the segmented partial point cloud of each strawberry. In our implementation, YOLOv8 and SAM2 require an average of 0.30\,s and 0.22\,s per frame, respectively, resulting in a total perception time well within the operational requirements of mobile harvesting robots.

The detected strawberries are ordered by the Euclidean distance between their estimated centers and the robot end-effector, and processed sequentially from nearest to farthest. This prioritization favors reachable targets and improves robustness under time and computational constraints.

\printcredits

\bibliographystyle{cas-model2-names}


\bibliography{references}



\Needspace{5cm}
\bio{ 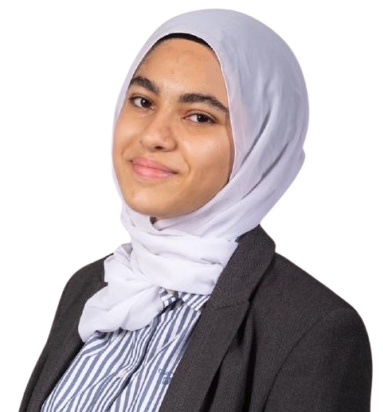}
\textbf{Malak Mansour} is currently pursuing her Ph.D. in Robotics at Mohamed bin Zayed University of
Artificial Intelligence (MBZUAI), Abu Dhabi,
United Arab Emirates. She also received her M.Sc. degree in Robotics from MBZUAI in 2026 and her B.Sc. degree in Electrical
Engineering with a minor in Computer Science from New York University Abu Dhabi (NYUAD), Abu Dhabi,
United Arab Emirates, in 2024. Her research interests include
robot perception and manipulation, uncertainty-aware robotic grasping, vision-language-action models, multilingual vision-and-language navigation, and robot-assisted surgery.
\endbio

\Needspace{5cm}

\bio{ 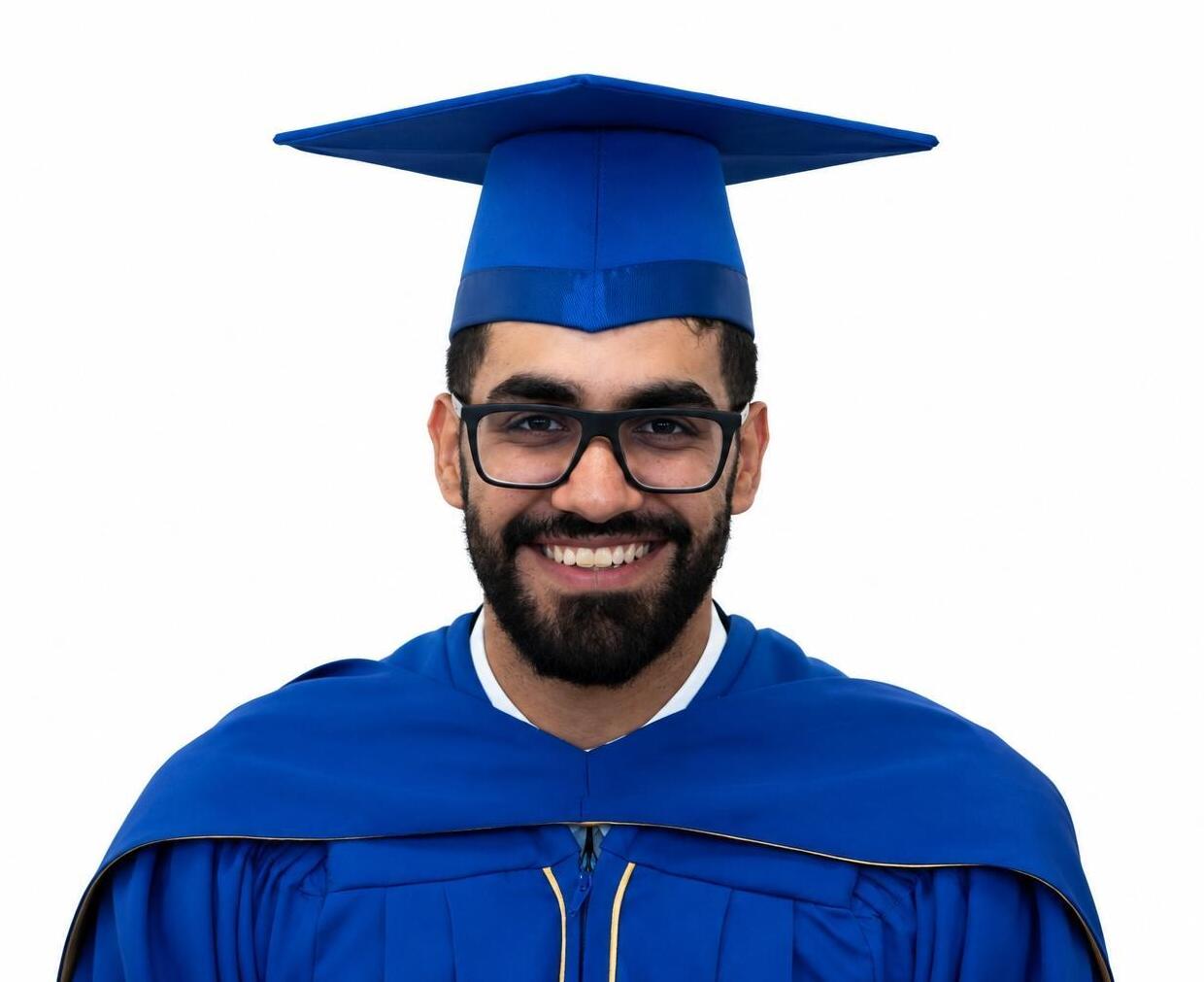}
\textbf{Ali Abouzeid}  is currently pursuing his Ph.D. in Robotics at Mohamed bin Zayed University of
Artificial Intelligence (MBZUAI), Abu Dhabi,
United Arab Emirates. He also received his M.Sc. degree in Robotics from MBZUAI in 2026. His research interests include robot
perception, three-dimensional vision, point-cloud completion,
robot learning, vision-language-action models, and robotic
manipulation. His recent work has investigated failure detection, geometry-aware
robot policies, perception, and grasp-planning systems.
\endbio

\Needspace{5cm}

\bio{ 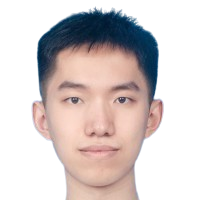}
\textbf{Zezhou Sun} is a Postdoctoral Researcher in the Department
of Robotics at the Mohamed bin Zayed University of Artificial
Intelligence (MBZUAI), Abu Dhabi, United Arab Emirates. His research
interests include simultaneous localization and mapping,
autonomous exploration, path planning, active loop closure,
robot perception, and robot learning. His work addresses reliable
robot navigation and manipulation in complex and unstructured
environments.
\endbio

\Needspace{5cm}

\bio{ 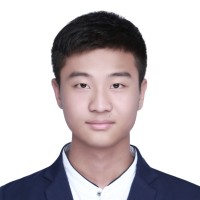}
\textbf{Qinbo Sun} is a Postdoctoral Researcher in the Department
of Robotics at the Mohamed bin Zayed University of Artificial
Intelligence (MBZUAI), Abu Dhabi, United Arab Emirates. His research
interests include robust autonomy in complex environments, marine
robotics, autonomous sailboats, energy-aware planning and control,
networked robotics, and long-duration robotic operation. His
research has appeared in journals and conferences including the
IEEE Transactions on Robotics, the Journal of Field Robotics,
IEEE Robotics and Automation Letters, ICRA, and IROS.
\endbio

\Needspace{5cm}

\bio{ 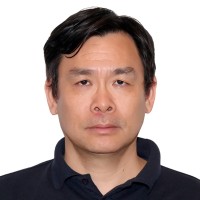}
\textbf{Dezhen Song} is the Vice Provost for Student and
Postdoctoral Affairs and a Professor of Robotics at the Mohamed
bin Zayed University of Artificial Intelligence (MBZUAI), Abu Dhabi,
United Arab Emirates. He received the Ph.D. degree in Operations
Research from the University of California, Berkeley, USA, and
the M.S. and B.S. degrees from Zhejiang University, China. His
research interests include robot spatial intelligence,
cross-modal perception and learning, robust navigation, scene
representation and understanding, and tightly coupled perception
and planning under uncertainty.
\endbio

\Needspace{5cm}

\bio{ 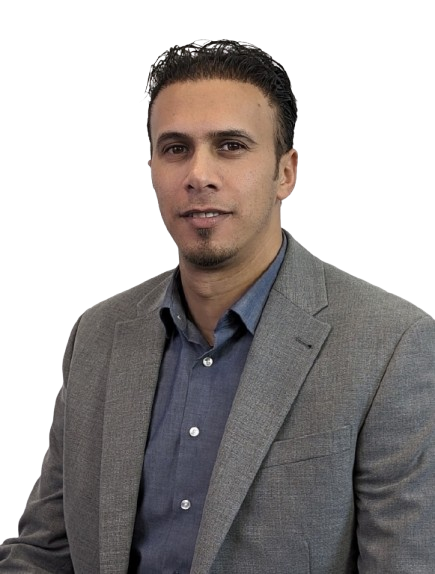}
\textbf{Abdalla Swikir} is an Assistant Professor of Robotics at
the Mohamed bin Zayed University of Artificial Intelligence (MBZUAI),
Abu Dhabi, United Arab Emirates. He received the B.Sc. degree in
Electrical Engineering from Omar Al-Mukhtar University, Libya,
the M.Sc. degree in Electrical Engineering from The Ohio State
University, USA, and the doctoral degree from the Technical
University of Munich, Germany. His research interests include
safe robot learning, control barrier functions, symbolic control,
robotic systems and mechatronics, and compositional analysis and
control of large-scale interconnected systems.
\endbio

\end{document}